%% file: paper.tex
\documentclass[]{bytedance_seed}

\usepackage[toc,page,header]{appendix}
\definecolor{dark-gray}{gray}{0.30}
\newcommand{\pub}[1]{{\color{dark-gray}{\scriptsize{[{#1}]}}}}
\usepackage{graphicx}
\usepackage{subcaption}
\usepackage{array}
\usepackage{multirow}    
\usepackage{xcolor}      
\usepackage{booktabs}    
\definecolor{tablegray}{RGB}{240,240,240}  
\definecolor{tableblue}{RGB}{230,245,255}  
\usepackage{multirow}    
\usepackage{xcolor}      
\usepackage{booktabs}    
\definecolor{tablegray}{RGB}{240,240,240}  
\definecolor{tableblue}{RGB}{230,245,255}  
\usepackage{booktabs} 
\usepackage{caption}  
\usepackage{multirow} 
\usepackage{svg}
\usepackage{mathtools}
\usepackage{bm}
\usepackage{amssymb}
\usepackage{capt-of}
\definecolor{iccvblue}{rgb}{0.21,0.49,0.74}
\PassOptionsToPackage{table,xcdraw}{xcolor}
\usepackage[utf8]{inputenc} 
\usepackage[T1]{fontenc}    
\usepackage{hyperref}       
\usepackage{url}            
\usepackage{booktabs}       
\usepackage{amsfonts}       
\usepackage{nicefrac}       
\usepackage{microtype}      
\usepackage{multirow}
\usepackage{tabularx}
\usepackage{multicol}
\usepackage{colortbl}
\usepackage{minitoc}
\usepackage{xcolor}      
\usepackage{xspace}
\usepackage{array}
\usepackage{textcomp}
\usepackage{multirow}    
\usepackage{xcolor}      
\usepackage{booktabs}    
\definecolor{tablegray}{RGB}{240,240,240}  
\definecolor{tableblue}{RGB}{230,245,255}  
\usepackage{multirow}    
\usepackage{xcolor}      
\usepackage{booktabs}    
\definecolor{tablegray}{RGB}{240,240,240}  
\definecolor{tableblue}{RGB}{230,245,255}  
\usepackage{booktabs} 
\usepackage{caption}  
\usepackage{multirow} 
\usepackage{xcolor}         
\usepackage[utf8]{inputenc} 
\usepackage[T1]{fontenc}    
\usepackage{hyperref}       
\usepackage{url}            
\usepackage{booktabs}       
\usepackage{amsfonts}       
\usepackage{nicefrac}       
\usepackage{microtype}      
\usepackage{listings}
\usepackage{xcolor} 
\usepackage{booktabs}
\usepackage{multirow}
\usepackage{makecell}
\usepackage{xcolor}

\usepackage{listings}
\usepackage{xcolor}          

\newcommand{\modelname}{UniUGP\xspace}

\title{UniUGP: Unifying Understanding, Generation, and Planing For End-to-end Autonomous Driving}

\author{ByteDance Seed}
\affiliation{See \hyperref[sec:contribution]{Contributions} section for a full author list.}

\abstract{
Autonomous driving (AD) systems struggle in long-tail scenarios due to limited world knowledge and weak visual dynamic modeling. Existing vision-language-action (VLA)-based methods cannot leverage unlabeled videos for visual causal learning, while world model-based methods lack reasoning capabilities from large language models. 
In this paper, we construct multiple specialized datasets providing reasoning and planning annotations for complex scenarios. Then, a unified Understanding-Generation-Planning framework, named \modelname, is proposed to synergize scene reasoning, future video generation, and trajectory planning through a hybrid expert architecture. By integrating pre-trained VLMs and video generation models, \modelname leverages visual dynamics and semantic reasoning to enhance planning performance. Taking multi-frame observations and language instructions as input, it produces interpretable chain-of-thought reasoning, 
physically consistent trajectories, and coherent future videos. We introduce a four-stage training strategy that progressively builds these capabilities across multiple existing AD datasets, along with the proposed specialized datasets. Experiments demonstrate state-of-the-art performance in perception, reasoning, and decision-making, with superior generalization to challenging long-tail situations.}

\checkdata[Project Page:]{\url{https://seed-uniugp.github.io/}}
\checkdata[Date:]{\today}

\begin{document}

\maketitle


\input{sections/introduction}

\input{sections/relatedwork}

\input{sections/approach}

\input{sections/experiments}

\clearpage

\bibliographystyle{plainnat}
\bibliography{main}

\clearpage

\beginappendix

\input{sections/appendix}

\end{document}

%% file: sections/introduction.tex


\begin{table*}[htbp]
  \centering
  \footnotesize 
  \renewcommand{\arraystretch}{1.2}  
  \caption{\textbf{Comparison of Different Methods.}  "VLA" refers to the use of additional models to predict more accurate trajectories, which is different from VLM's text-based prediction. "Reason." refers to whether the model can generate a chain of thoughts. "Inter." refers to whether the model can change its trajectory based on human instructions. "Cont." refers to whether the token is a continuous or discrete value. "World Model" only lists the methods that can simultaneously generate future images and trajectories. "FM." means flow matching.}
  \begin{tabularx}{\linewidth}{llcXXXXXc }  
    \toprule
      &
      &
     &
    \multicolumn{2}{c}{\cellcolor{tablegray}\bfseries Understanding} &
    \multicolumn{2}{c}{\cellcolor{tablegray}\bfseries Generation} &
    \multicolumn{2}{c}{\cellcolor{tablegray}\bfseries Action} \\
    \cmidrule(lr){4-5} \cmidrule(lr){6-7} \cmidrule(lr){8-9}  
    \cellcolor{tablegray}\centering\bfseries Category &
    \cellcolor{tablegray}\centering\bfseries Method &
    \cellcolor{tablegray}\centering\bfseries Model &
    \cellcolor{tablegray}\bfseries Reason. &
    \cellcolor{tablegray}\bfseries Inter. &
    \cellcolor{tablegray}\bfseries Modality &
    \cellcolor{tablegray}\bfseries Cont. &
    \cellcolor{tablegray}\bfseries Method &
    \cellcolor{tablegray}\bfseries Cont. \\
    \midrule
    World Model & OccWorld~\cite{zheng2023occworld} & - & $\times$ & $\times$ & Occ & $\times$ & Codebook & $\times$ \\  
    World Model & Epona~\cite{Zhang2025EponaAD} & - & $\times$ & $\times$ & Video & $\surd$ & FM. & $\surd$ \\
    \cmidrule(lr){1-9}
    VLM & DriveLM~\cite{sima2023drivelm} & Llama2 & $\times$ & $\times$ & - & - & Text & $\times$ \\
    VLM & Impr. VLA~\cite{Chi2025ImpromptuVO} & Qwen2.5-VL-3B & $\times$ & $\times$ & - & - & Text & $\times$ \\
    VLM & OmniDrive~\cite{Wang2024OmniDriveAH} & LLaMA2-7B & $\times$ & $\times$ & - & - & Text & $\times$ \\
    \cmidrule(lr){1-9}
    VLA & ReCogDrive~\cite{Li2025ReCogDriveAR} & Qwen2.5-VL-3B & $\surd$ & $\times$ & - & - & Diffusion & $\surd$ \\
    VLA & ORION~\cite{Fu2025ORIONAH} & Vicuna v1.5 & $\times$ & $\times$ & - & - & VAE & $\times$ \\
    VLA & AutoVLA~\cite{Zhou2025AutoVLAAV} & InternVL3-8B & $\surd$ & $\times$ & - & - & Codebook & $\times$ \\
    VLA & DriveMoE~\cite{yang2025drivemoe} & $\pi$0 & $\times$ & $\times$ & - & - & FM. & $\surd$ \\
    VLA & Alpamayo-R1~\cite{wang2025alpamayo} & Cosmos-Reason & $\surd$ & $\times$ & - & - & FM. & $\surd$ \\
    \cmidrule(lr){1-9}
    Unified Model & Doe-1~\cite{Zheng2024Doe1CA} & Lumina-mGPT & $\times$ & $\times$ & Video & $\times$ & Text & $\times$ \\
    Unified Model & Occ-LLM~\cite{xu2025occ} & Llava-7B & $\times$ & $\times$ & Occ & $\times$ & Text & $\times$ \\
    Unified Model & OccLlama~\cite{wei2024occllama} & Llama-7B & $\times$ & $\times$ & Occ & $\times$ & Text & $\times$ \\
    Unified Model & HERMES~\cite{Zhou2025HERMESAU} & InternVL2-2B & $\times$ & $\times$ & LiDAR & $\times$ & Text & $\times$ \\
    Unified Model & FSDrive~\cite{zeng2025futuresightdrive} & Qwen2-VL-2B & $\times$ & $\times$ & Video & $\times$ & Text & $\times$ \\
    \cmidrule(lr){1-9}
    \rowcolor{tableblue}
    Unified Model &  \modelname &   Qwen2.5-VL-3B &   $\surd$ &  $\surd$ &  Video &   $\surd$ &  FM. &  $\surd$ \\
    \bottomrule
  \end{tabularx}
  \label{tab:method_comparison}
\end{table*}

\section{Introduction}
\label{sec:intro}
Autonomous driving (AD) has recently made remarkable progress, especially in areas such as bird's-eye view perception~\cite{li2024bevformer,huang2021bevdet,liu2022petr,lu2023towards,lu2024scaling}, end-to-end~\cite{hu2023_uniad,diffusiondrive,jiang2023vad,chen2024vadv2}, scene reconstruction~\cite{zhou2024drivinggaussian,yan2024street,zhao2025drivedreamer4d,lu2024drivingrecon}, and video generation~\cite{gao2023magicdrive,wang2024drivedreamer,wen2024panacea,lu20254d}. Recently, given the superior capabilities of multimodal large language models (MLLMs) in world knowledge, reasoning ability, and interpretability, they have been widely applied in AD~\cite{hu2025contextalignment,ma2023dolphins,zhang2024chatscene,li2024llada}. One promising direction is the end-to-end vision-language-action (VLA) model~\cite{Zhou2025AutoVLAAV,wang2025alpamayo,Li2025ReCogDriveAR}, which leverages pre-trained vision-language model (VLM) to directly extract scene features from visual observations and language instructions, subsequently generating vehicle control commands (e.g., speed and trajectory). This paradigm not only simplifies system architecture and minimizes information loss, but also enables the utilization of the model's world knowledge to analyze driving environments and reason about safe decisions in complex scenarios~\cite{Cai2024DrivingWR, Jiang2025AlphaDriveUT, Jiang2024SennaBL, Wang2023DriveMLMAM}. However, they were unable to fully utilize the large number of un-labeled driving videos, which limited their ability to learn visual causal reasoning from large-scale datasets.

In addition to the advanced VLA technology, the world model can learn visual causal reasoning by predicting the next frame of the video~\cite{Zhang2025EponaAD,zheng2023occworld,hu2024drivingworld,yang2024generalized,Chen2024DrivingGPTUD}. The world model can learn visual causal reasoning by predicting the next frame of the video, which has been proven to be helpful in achieving the final end-to-end AD~\cite{Zhang2025EponaAD,yang2024generalized,zeng2025futuresightdrive}. But the world model is unable to match the world knowledge, reasoning ability, and interaction capability from large language models. Summaries of different methods are presented in Tab.~\ref{tab:method_comparison}. Unified models, which aim to bridge perception, reasoning, and action, can simultaneously combine the advantages of the world model and the VLA model. However, there are several additional issues here: 1) How to efficiently establish a unified model to fully utilize the pre-trained VLM and world model. 2) How to effectively and repeatedly utilize various driving data (such as VQA pairs, video trajectory pairs, etc.) to fully exploit the potential of the unified model. 3) How to evaluate the capabilities of the unified model, especially in terms of understanding, reasoning, and planning in complex scenarios.

To address these limitations, we propose \modelname, a unified Understanding–Generate–Planning framework for end-to-end AD that jointly models complex scene reasoning, future video generation, and trajectory planning. Built on a hybrid expert architecture, \modelname fully exploits the causal reasoning capabilities of pre-trained MLLMs and video generation models, while further enhancing cross-modal causal alignment through large-scale multimodal data training. Specifically, \modelname takes natural language instructions and continuous image sequences as inputs, and outputs three complementary results: a chain-of-thought (CoT) reasoning process for interpretability, a physically consistent trajectory for safe driving, and a coherent future video for visual causal validation. To ensure output consistency and accuracy, we design a multi-term loss function that enforces CoT logical consistency, trajectory temporal smoothness, and video visual coherence. Moreover, we propose a four-stage training framework that sequentially builds foundational scene understanding, visual dynamic modeling, text-based reasoning, and multi-capability fusion, leveraging over 10 diverse AD datasets to cover common scenarios and long-tailed cases.

Our main contributions are summarized as follows:

1) We construct multiple specialized datasets for AD-oriented VLAs, which provides explanations, reasoning, and planning for complex scenarios.

2) We propose a unified Understanding-Generation-Planning (\modelname) framework with a hybrid expert architecture, which synergizes scene reasoning, future video generation, and trajectory planning.


3) We develop a four-stage training strategy that leverages diverse AD datasets to enable the mutual enhancement of understanding, generation, and planning task capabilities.


%% file: sections/relatedwork.tex
\section{Related Work}
\label{sec:related}

\subsection{VLA for Autonomous Driving}

End-to-end autonomous driving models ~\cite{Hu2022UniAD, jiang2023vad,yang2024genad,Liao2024DiffusionDriveTD} have shown strong performance in structured environments but struggle in long-tail and unstructured scenarios due to limited generalization and lack of world knowledge. To mitigate this, recent work introduces VLMs for reasoning and scene understanding ~\cite{Bai2025Qwen25VLTR,Qwen2-VL}, integrating them into driving frameworks to enhance adaptability in complex situations ~\cite{Zhou2025AutoVLAAV, Li2025ReCogDriveAR, Chi2025ImpromptuVO, Jiang2025DiffVLAVG, Fu2025ORIONAH, Wang2024OmniDriveAH}. Early attempts used VLMs to generate high-level meta-actions or abstract driving decisions ~\cite{Cai2024DrivingWR, Jiang2025AlphaDriveUT, Jiang2024SennaBL, Wang2023DriveMLMAM}, which guided modular or end-to-end planners but disrupted joint optimization across perception, decision, and control. Later, the VLA frameworks directly mapped visual language inputs to trajectories. Impromptu VLA ~\cite{Chi2025ImpromptuVO} trained VLMs on text-based trajectory representations, while AutoVLA ~\cite{Zhou2025AutoVLAAV} discretized trajectories into action tokens decoded into continuous paths. Diffusion-based models such as ReCogDrive, DiffVLA, and ORION ~\cite{Li2025ReCogDriveAR, Jiang2025DiffVLAVG, Fu2025ORIONAH} further bridged the gap between semantic reasoning and continuous trajectory generation. Despite progress, the existing methods are often unable to utilize unlabeled video data to learn visual causal reasoning.

\subsection{World Models for Autonomous Driving}

World models~\cite{zhao2024DriveDreamer4D, Ni2024ReconDreamerCW, zuo2024gaussianworld, bian2025dynamiccity, Xu2025OccLLMEA} aim to infer ego-centric states and predict dynamic surroundings from historical observations, enabling accurate future prediction and planning. In autonomous driving, world models are primarily applied in three areas: driving scenario generation~\cite{Ni2025MaskGWMAG, li2024uniscene, Hassan2024GEMAG, li2025semi}, planning~\cite{Hu2022ModelBasedIL, wang2023driving, zheng2023occworld}, and representation learning~\cite{Min2024DriveWorld4P, Yang_2024_CVPR, zeng2024driving}. 
For driving scenario generation, most prior works rely on diffusion models. GAIA-1~\cite{Hu2023GAIA1AG} is a notable exception that combines a progressive next-token predictor with an auxiliary diffusion image decoder. More recently, Epona~\cite{Zhang2025EponaAD} advances this direction by employing an autoregressive diffusion model to unify world modeling and planning. 
However, existing approaches face two major limitations. First, they do not fully exploit the complementary strengths of pre-trained VLMs and video generation models, missing opportunities to leverage both linguistic reasoning and visual dynamics modeling. Second, most methods are trained and evaluated on limited single-dataset scenarios, restricting their generalization ability. Our method addresses these gaps by integrating knowledge from both pre-trained VLMs and video generative models, while scaling training across diverse datasets to unlock emergent capabilities.

\subsection{Unified Models}
Unified models aim to dissolve modular boundaries across understanding and generation. ~\cite{Deng2025bagel,shi2024lmfusion, pan2025transfer, sun2023emu, team2024chameleon, tong2024metamorph, wang2024emu3, wu2025janus, ma2025janusflow, xie2024show, xie2025show, zhou2024transfusion}.  Extending to embodied intelligence, works like ~\cite{Lv2025F1AV} and WALL-OSS ~\cite{Zhai2025walloss} integrate action modules into unified frameworks. F1 uses a three-expert MoT to synthesize goal-conditioned visual foresight and model action as foresight-guided inverse dynamics, while WALL-OSS employs a tightly coupled MoE design to align discrete action priors and continuous control, enhancing long-horizon manipulation. 
In autonomous driving, recent works~\cite{Zheng2024Doe1CA,Zhou2025HERMESAU,zeng2025futuresightdrive} have explored unified architectures. However, as shown in Tab.~\ref{tab:method_comparison}, they lack critical capabilities including CoT reasoning, natural language interaction, and diffusion-based planning, while failing to leverage large-scale unlabeled data, and pre-trained VLM and video generation model knowledge—limiting both their interpretability and generalization potential.

%% file: sections/approach.tex
\section{Method}

The proposed \modelname is a unified model for understanding, generation, and planning to further enhance the causal reasoning ability across different modalities: 1) Large-scale challenging data pairs are collected and processed to train and evaluate the understanding, reasoning, generation and planning capabilities of the unified model, as elaborated in Sec.~\ref{sec:data}. 2) An elegant unified framework combines the advantages of VLA and world models, as elaborated in Sec.~\ref{sec:framework}. 3) A well-designed training strategy enables the unified model to fully acquire knowledge from different modalities on various datasets, as elaborated in Sec.~\ref{sec:train}.

\begin{figure*}[t]
    \centering
    \includegraphics[width=0.8\textwidth]{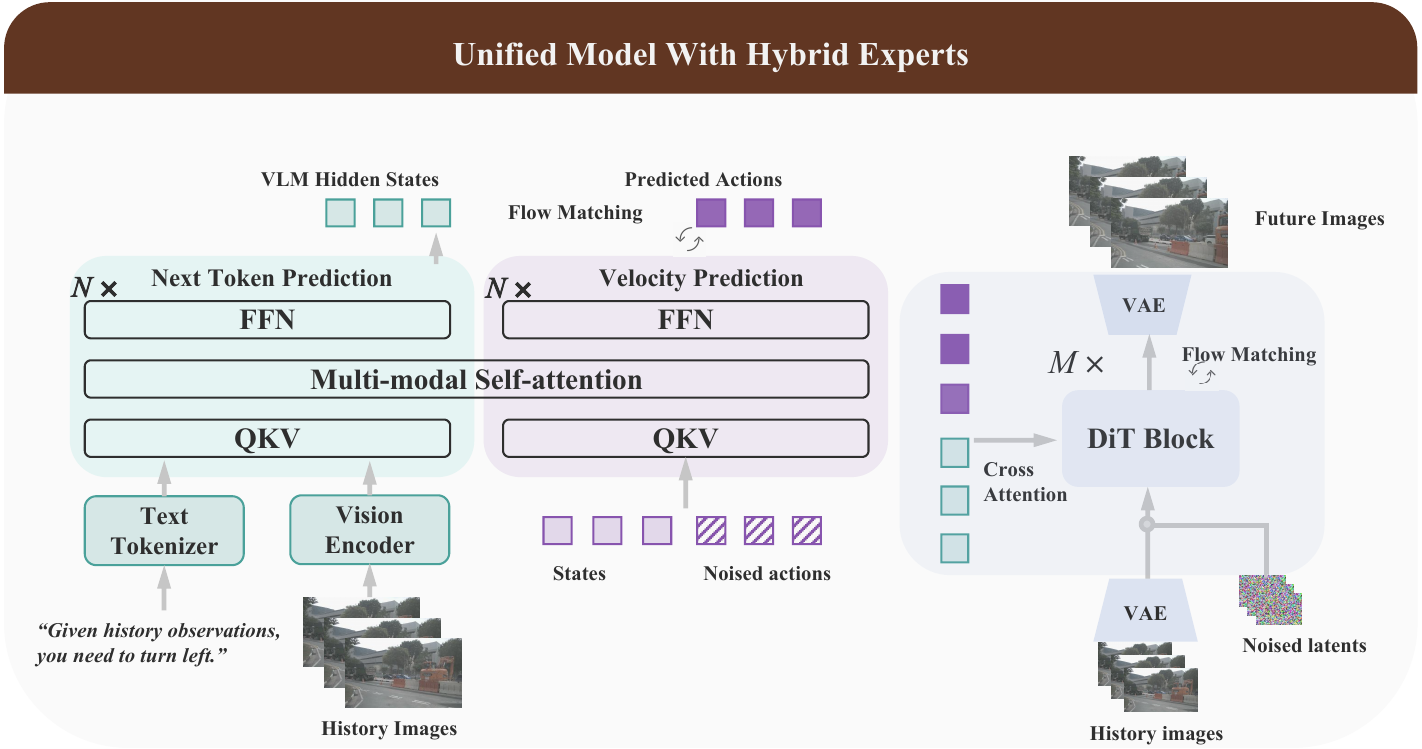}
    \caption{\textbf{Illustration of UniUGP}, a unified model with three hybrid 
    experts. The understanding expert performs the next-token prediction for causal reasoning. The planning expert forms a MoT architecture with the understanding expert, and performs the velocity prediction in flow matching for production future actions. The generation expert is cascaded as a world model to produce future videos.} 
    \label{fig:method}
\end{figure*}

\begin{figure*}[htbp]
    \centering
    \includegraphics[width=0.95\textwidth]{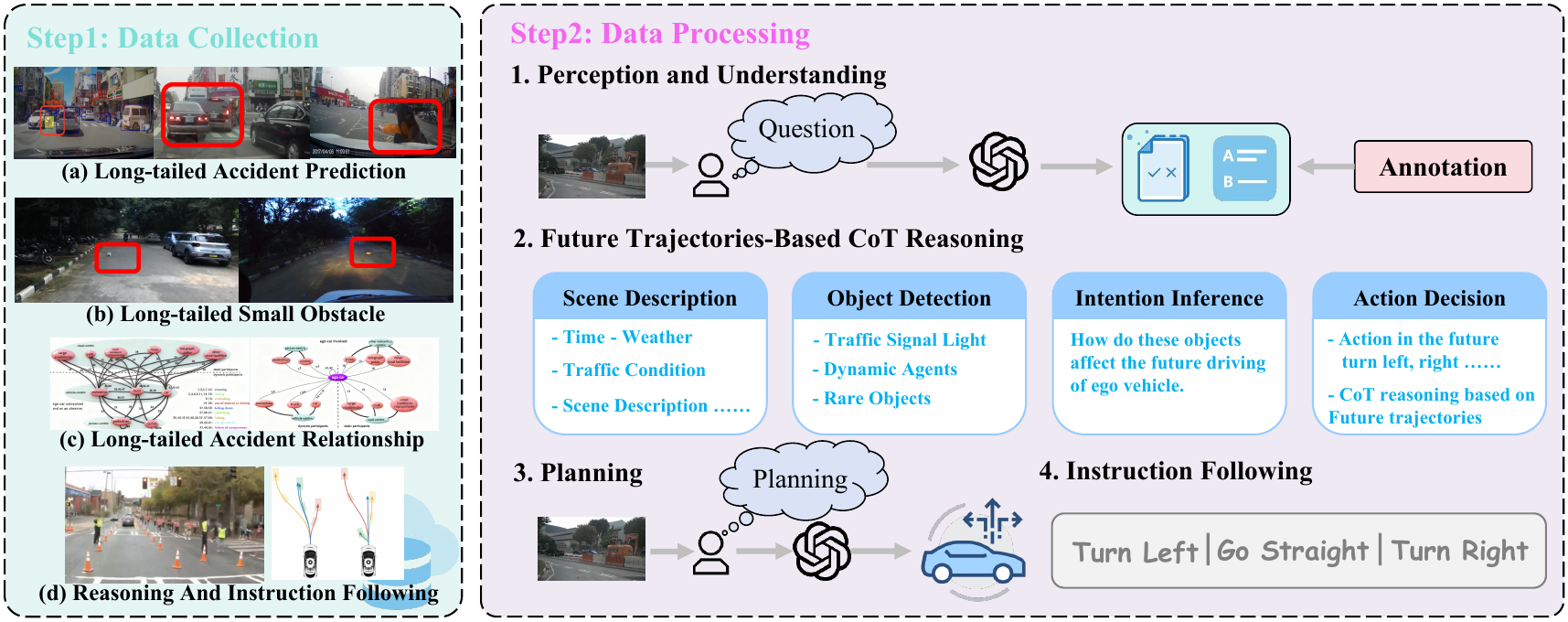}
    \caption{\textbf{Dataset Construction Pipeline.} This figure depicts the pipeline of data collection (integrating multiple challenging driving datasets) and data processing (featuring four task categories: understanding,  chain-of-thought, planning, and instruction following) to train and assess the cognitive abilities of end-to-end autonomous driving models within a unified QA framework.} 
    \label{fig:method}
\end{figure*}

\subsection{Challenging Long-tail Driving Dataset}
\label{sec:data}
Prior benchmarks emphasize structured scenes and simulator-based evaluations~\cite{DriveVLM,Wang2024OmniDriveAH,Chi2025ImpromptuVO}, which overlook the challenges of long-tail driving events. To address this, we have collected a large number of challenging long-tail driving videos, including Waymo-E2E~\cite{xu2025wod}, DADA2000 ~\cite{9312486}, Lost and Found (LaF) ~\cite{pinggera2016lost}, StreetHazards (StHa)~\cite{hendrycks2019anomalyseg}, SOM~\cite{singh2020lidar}, and AADV ~\cite{chan2016anticipating}.

These datasets were further processed to be used separately for training and evaluating the perception ability, causal reasoning ability, planning ability, and the ability to follow instructions. For the comprehension task, we have designed three sub-tasks: small objects, accident subject relationships, and accident prediction. The questions and answers of these tasks are labeled based on the provided labels from the dataset and the advanced VLM model. For more details, please refer to the supplementary materials. CoT reasoning: We employed the results of future planning and reasonable prompts to force the advanced VLM to generate the accurate CoT. We carried out manual calibration for CoT. Finally, it is worth mentioning that we will assign corresponding instructions to each predicted trajectory in the future, which enables our model to have the ability to follow instructions. The more detailed data processing procedures are listed in the Appendix.




\subsection{Unified Model With Hybrid Expert}
\label{sec:framework}
We adopted the Hybrid Expert structure to achieve unified understanding, generation and planning. This architecture can leverage the advanced features of the diffusion policy and the knowledge of pre-trained VLM and world models. Unified Model

\noindent\textbf{Understanding and Planning Experts.} Based on the validity proof of the existing work~\cite{Fu2025ORIONAH,black2410pi0,Lv2025F1AV,shi2024lmfusion}, the understanding and planning experts constitute a Mixture-of-Transformers (MoT) architecture, as shown in Figure \ref{fig:method}. For the understanding expert, we choose the Qwen2.5-VL~\cite{Qwen2.5-VL} as our backbone model. The text instructions and the observation images are firstly mapped to aligned cross-modal understanding tokens $\bm{x}^{{und}}$, by the text tokenizer and the ViT encoder, respectively. 

For the planning expert, the action chunk $\bm{a}$ is modeled by a flow matching process \cite{lipman2022flow}, where the expert learns to reverse a gradual noise-addition process added in the forward process. During training, a random noise $\epsilon \sim \mathcal{N}(\bf{0}, \bf{I})$ and a timestep $\tau \in [0, 1]$ are sampled to model the noised actions:
\begin{equation}
\bm{a}_{\tau} = \tau \bm{a} + (1 - \tau) \epsilon 
\end{equation}
The $\bm{a}_{\tau}$ along with the history states $\mathbf{s}$ are prjoected into planning expert tokens $\bm{x}^{plan} = \mathrm{Proj.} ([\bm{s}, \bm{a}_{\tau}])$.

The $\bm{x}^{{und}}$ and $\bm{x}^{plan}$ are passed through several MoT layers, where each layer can be formalized as:
\begin{equation}
\bm{h}_{o}^{und}, \bm{h}_{o}^{und} = \mathrm{MSHA} ([\mathrm{QKV}_{und}(\bm{x}^{{und}}), \mathrm{QKV}_{plan}(\bm{x}^{{plan}})])
\end{equation}
The $\mathrm{QKV}$ denote the linear projections that map understanding and planning tokens (or hidden states) into queries, keys, and values. The MSHA denotes the multi-head self-attention. Then, the modality-specific feed-forward networks (FFNs) are utilized to process the $\bm{h}_{o}^{und}$ and $\bm{h}_{o}^{und}$, separately.
\begin{align}
\bm{h}_{ffn}^{und} = \mathrm{FFN}_{und}(\bm{h}_{o}^{und}), \\
\bm{h}_{ffn}^{plan} = \mathrm{FFN}_{plan}(\bm{h}_{o}^{plan})
\end{align}
The final resulting hidden states $\bm{h}^{und}$ and $\bm{h}^{plan}$ are mapped into the logits in text and the predicted denoising vector filed.
\begin{align}
P_{logits} = & \mathrm{LMHead}(\bm{h}^{und}), \\
\bm{u}_{\tau}^{plan} = & \mathrm{unProj.}(\bm{h}^{plan})
\end{align}
The training objectives for the understanding and planning experts are formalized as:
\begin{align}
\mathcal{L}_{und} = & \mathbb{E}_{{x}_i^{{und}}}[-\mathrm{log}(P({x}_i^{{und}} | \bm{x}_{<i}^{{und}}))], \\
\mathcal{L}_{plan} = & \mathbb{E}_{\bm{u}_{\tau}^{plan}}[|| \bm{u}_{\tau}^{plan} - (\epsilon - \bm{a}) ||_2]
\end{align}

\noindent\textbf{Generation Expert.} 
The generation expert interacts with the understanding and planning experts in a serial manner. On mobile devices, the generation expert can be disabled to save computational effort, without compromising the performances of the former experts.

The generation expert produces future videos via a flow matching process, same as the planning expert. It is composed of several DiT \cite{peebles2023scalable} blocks. In practice, we adopt Wan2.1 \cite{wan2025wan} as the base model and inherit its pre-trained parameters. Note that any other DiT-based video generation models are feasible here. 

Both future and history images are encoded into tokens by VAE. As shown in Figure \ref{fig:method}, the history image tokens $\bm{v}^{hist}$ and noised future image tokens $\bm{v}_{\tau}^{fut}$ are concatenated as the input. The understanding hidden states $\bm{h}^{und}$ and the action embeddings $\bm{A}$ are concatenated as the condition. The final denoising vector feild is computed as:
\begin{align}
\bm{u}_{\tau}^{gen} = \mathcal{W}([\bm{v}^{hist}, \bm{v}_{\tau}^{fut}], [\bm{h}^{und}, \bm{A}], \tau)
\end{align}
During inference, the action embeddings are obtained by projecing the predicted future actions $\hat{\bm{a}}$ from the planning expert, as $\bm{A}=\mathrm{Proj.}(\hat{\bm{a}})$. During training, The embeddings are computed either from ground truth actions, or from actions obtained via single-step denoising:
\begin{equation}
\bm{A} \sim 
\begin{cases} \mathrm{Proj.}(\bm{a}) & \text{if rand} (0 \sim 1) > 0.5, \\
 \mathrm{Proj.} (\bm{a}_{\tau} - (1-\tau)\bm{u}_{\tau}^{plan}) & \text{else}.
\end{cases}
\end{equation}
As such, the generation expert receives both semantic and physical signals from the understanding and planning experts, facilitating more realistic video generation. The training objective can be formalized as:
\begin{align}
\mathcal{L}_{gen} = & \mathbb{E}_{\bm{u}_{\tau}^{gen}}[|| \bm{u}_{\tau}^{gen} - (\epsilon - \bm{v^{fut}}) ||_2]
\end{align}

\subsection{Training Recipe}

\begin{table*}[t]
\centering
\caption{Hyperparameter Configuration for the Four-Stage Training Framework}
\label{tab:hyperparameters}
\renewcommand{\arraystretch}{1.15}
\setlength{\tabcolsep}{6pt}
\begin{tabularx}{\linewidth}{l *{4}{X}}
\toprule
\textbf{Hyperparameter} & \textbf{Stage 1} & \textbf{Stage 2} & \textbf{Stage 3} & \textbf{Stage 4} \\
\midrule

\multirow{3}{*}{\textbf{Trained Components}}
 & Und. Expert: \checkmark & Und. Expert: $\times$ & Und. Expert: \checkmark & Und. Expert: \checkmark \\
 & Gen. Expert: $\times$ & Gen. Expert: \checkmark & Gen. Expert: $\times$ & Gen. Expert: \checkmark \\
 & Plan. Expert: $\times$ & Plan. Expert: \checkmark & Plan. Expert: $\times$ & Plan. Expert: \checkmark \\

\midrule

\textbf{Dataset}
 & Custom long-tail dataset, ImpromptuVLA \cite{Chi2025ImpromptuVO}
 & nuScenes \cite{Caesar2019nuScenesAM}, NuPlan \cite{caesar2021nuplan}, Waymo \cite{waymo}, Lyft \cite{li2023large}, Cosmos \cite{nvidia2025cosmosdrivedreams}
 & Custom CoT dataset
 & Mixture of Stages 1--3 (ratio 0.1 : 0.4 : 0.5) \\

\midrule

\textbf{Learning Rate}
 & $10^{-4}$ & $10^{-4}$ & $10^{-4}$ & $10^{-4}$ \\

\midrule

\textbf{Understanding Resolution}
 & (224, 224) & -- & (224, 224) & (224, 224) \\

\midrule

\textbf{Generation Resolution}
 & -- & (512, 512) & -- & (512, 512) \\

\midrule

\textbf{GPU Resources}
 & 8 nodes $\times$ 8 $\times$ gpus (80GB)
 & 8 nodes $\times$ 8 $\times$ gpus (80GB)
 & 8 nodes $\times$ 8 $\times$ gpus (80GB)
 & 8 nodes $\times$ 8 $\times$ gpus (80GB) \\

\midrule

\textbf{Batch Size}
 & 64 & 64 & 64 & 64 \\

\midrule

\textbf{Training Steps}
 & 1M & 4M & 1M & 4M \\



\bottomrule
\end{tabularx}
\end{table*}

\label{sec:train}
We design a four-stage training framework, which sequentially builds foundational scenario understanding, visual dynamic modeling, text-based reasoning, and multi-capability fusion, detailed are presented below. The training parameters and details can be found in  Tab.~\ref{tab:hyperparameters}.

\noindent\textbf{Stage 1: Continuous Training for the Understanding}  of the basic scenarios is to enable the Understanding Expert to establish a comprehensive understanding of diverse driving scenarios—covering common traffic and long-tailed cases. Only the Understanding Expert is trained in this stage. The training dataset at this stage includes the long-tail data set that we have labeled and the ImpromptuVLA (80,000 meticulously curated video clips from 8 open-source large-scale datasets)~\cite{Chi2025ImpromptuVO}.

\noindent\textbf{Stage 2: Visual Dynamics Modeling and Planning Training} focuses on learning visual dynamics and motion planning capabilities. During this training phase, driving videos with trajectories were used for training the Generation Expert and the Planning Expert. We utilized multiple public datasets including: nuScenes~\cite{Caesar2019nuScenesAM}, NuPlan~\cite{caesar2021nuplan}, Waymo~\cite{waymo}, Lyft~\cite{li2023large}, and Cosmos~\cite{nvidia2025cosmosdrivedreams}.

\noindent\textbf{Stage 3: Text Reasoning Learning for Causal Validation}  integrates CoT reasoning into the Understanding Expert, enabling the model to validate the logic of its perceptions and planning using natural language. This stage enhances model interpretability and ensures decisions are grounded in explicit causal reasoning.  This part is trained using our own annotated CoT dataset.

\noindent\textbf{Stage 4: Mixed Training for Multi-Capability Fusion} resolves potential misalignments between individual stages and enhances generalization across scenarios. In Stage 4, the three experts are jointly trained to achieve consistent end-to-end performance. We mix datasets from Stages 1–3 at a fixed proportion to balance foundational, reasoning, planning, and generation capabilities. The overall objective $\mathcal{L}_{\text{total}}$ is a weighted sum of three sub-objectives:  
\begin{equation}
\mathcal{L}_{{total}} = \alpha \cdot \mathcal{L}_{{und}} + \beta \cdot \mathcal{L}_{{plan}} + \gamma \cdot \mathcal{L}_{{gen}}
\end{equation}
where $\alpha=0.3$, $\beta=0.5$, $\gamma=0.2$.
This alignment ensures the model operates as a unified system rather than a collection of isolated components.

%% file: sections/experiments.tex
\section{Experiment}

\subsection{Dataset} 

We utilize two complementary categories of datasets to evaluate multimodal world understanding and decision-making under both common and safety-critical driving conditions. For Perception and Understanding, we adopt anomaly and accident anticipation datasets including DADA2000 ~\cite{9312486}, Lost and Found (LaF) ~\cite{pinggera2016lost}, StreetHazards (StHa)~\cite{hendrycks2019anomalyseg}, SOM~\cite{singh2020lidar}, and AADV ~\cite{chan2016anticipating}. These datasets contain rare obstacles, out-of-distribution objects, and pre-accident sequences collected in diverse real-world environments, enabling assessment of visual scene comprehension, hazard awareness, and recognition of long-tail semantic patterns.

For reasoning, planning, and instruction following, we use the Waymo Open Dataset Long-tail End-to-End Driving~\cite{xu2025wod}, which consists of 4,021 real-world driving segments specifically curated to capture rare and high-risk events that occur in less than 0.003\% of daily driving. Each segment provides surround-view camera streams, ego-motion history, and route signals, and the benchmark task is to predict future 5-second trajectories under uncertain and interaction-heavy conditions. This dataset enables rigorous evaluation of robustness, generalization, and physically grounded decision-making in long-tail scenarios, complementing the perception-focused anomaly datasets above.

Following the previous methods~\cite{cube-llm,Wang2024OmniDriveAH}, we evaluate scene understanding on DriveLM~\cite{sima2023drivelm}. This dataset features keyframe descriptions paired with QA annotations covering full-stack autonomous driving (perception, prediction, planning), offering comprehensive language support for development. 

Consistent with methods~\cite{jiang2023vad,gao2024vista}, we evaluate trajectory planning and future frames generation on the nuScenes~\cite{Caesar2019nuScenesAM}. The nuScenes contains 1,000 scenes of approximately 20 seconds each captured by a 32-beam LiDAR and six cameras providing 360-degree field of view. Specifically, the dataset provides 28,130 (train), 6,019 (val), and 193,082 (unannotated) samples. In addition, we utilize a large number of publicly available datasets for training, including: ImpromptuVLA~\cite{Chi2025ImpromptuVO}, NuPlan~\cite{caesar2021nuplan}, Waymo~\cite{waymo}, Lyft~\cite{li2023large}, and Cosmos~\cite{nvidia2025cosmosdrivedreams}.

\subsection{Metrics} We have established a new long-tail benchmark consisting of understanding, CoT reasoning, Planning and Instruction Following. For comprehension questions (mainly multiple-choice and true/false questions), we evaluate them based on the accuracy rate. For CoT, we use the API of GPT-4o to rate it in terms of consistency, rationality and fluency, and also provide a score from Blue. For Planning, we use L2 3s to rate it. For instruction following, we score the trajectories corresponding to different instructions using L2 (3s). More details are provided in the appendix. We evaluate trajectory planning using L2 displacement error and collision rate following previous methods~\cite{Zhang2025EponaAD}. Following existing methods~\cite{wang2023drivedreamer,yang2024genad}, we report Fréchet Inception Distance (FID)~\cite{fid} to measure the future frames generation quality. DriveLM GVQA~\cite{sima2023drivelm} metrics include language metrics like BLEU, ROUGE\_L, and CIDEr for text generation, the ChatGPT Score for open-ended Q\&A and accuracy for multiple-choice questions.


\subsection{Evaluation of Understanding Ability}

To validate the effectiveness of our benchmark and the performance of our model, we conduct comparative experiments with state-of-the-art vision-language models, and the results are presented in Tab.~\ref{tab:driving_performance}. The evaluation metrics are categorized into four key dimensions to comprehensively measure model performance: Understanding (assessing scene and object comprehension, including Small (small object recognition), accident subject relationship, and Acci.Pred. (accident event prediction)), CoT (Chain-of-Thought reasoning ability, evaluated by GPT (subjective GPT score) and Blue (BLEU score)), Planning (short-term driving planning, measured by L2 distance (3s) with smaller values indicating better performance), and Following (trajectory following accuracy, also measured by L2 distance (3s)). The comparative models include GPT 4o, Qwen 2.5 VL, and two ablated versions of our model (Our w/o CoT: without Chain-of-Thought module; Our w/o Gen.: without generation module), alongside our full model (Our).

\begin{figure*}[t]
    \centering
    \includegraphics[width=0.9\textwidth]{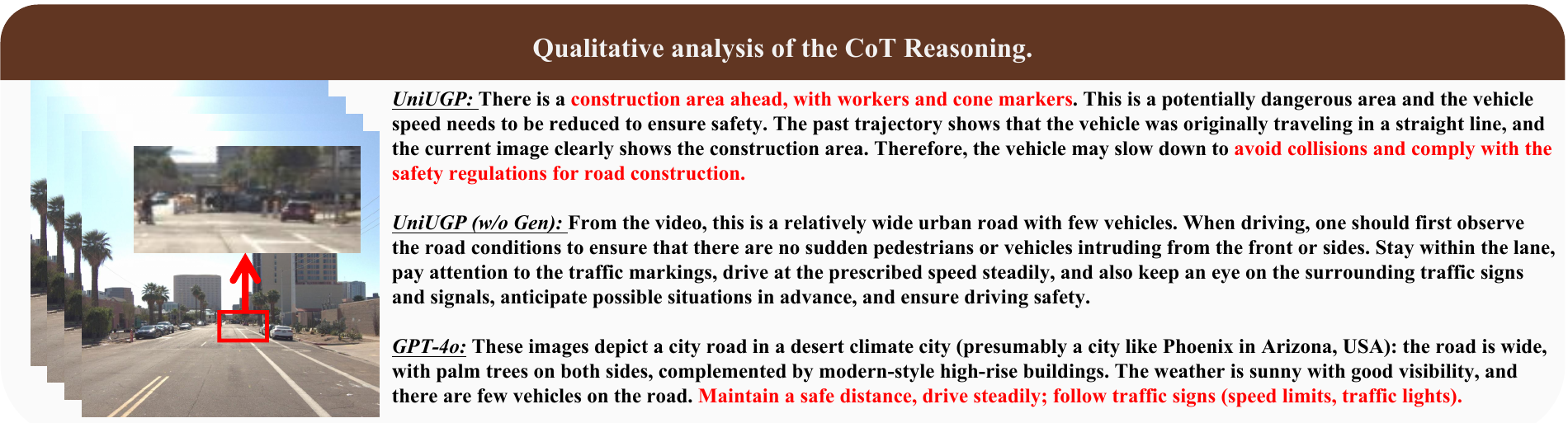}
    \caption{\textbf{The ablation experiment on the absence or presence of world model knowledge.} The world model enables the VLA to pay more attention to future causal relationships, thereby focusing on the semantics of distant objects. } 
    \label{fig:qf}
\end{figure*}

As shown in Table \ref{tab:driving_performance}, our full model outperforms state-of-the-art methods (GPT-4o, Qwen-2.5-VL-72B) and our ablated versions (Our w/o CoT, Our w/o Gen.) across all key metrics: in Understanding (89.3\%, 88.6\%, 95.8\% for Small, Relationship, Abnor. Pred.), CoT (GPT score 0.88, Blue score 0.240), Planning (L2=1.45), and Following (L2=1.40), all surpassing baselines and ablated models with lower performance; these results confirm that integrating the generation and CoT modules significantly enhances the model’s comprehensive driving capabilities. The generation model has significantly enhanced the performance of the model. We conducted a further qualitative analysis as shown in Fig.~\ref{fig:qf}. The world model forces VLA to learn visual causal inference, particularly focusing on distant objects to generate better future frames. This enables the VLA model to predict potential dangers in advance, thereby ensuring driving safety.

\begin{table*}[t] 
\centering
\caption{\textbf{Performance Comparison on Driving Evaluation Benchmark.} Note: GPT-4o and Qwen-2.5-VL-72B are provided with historical trajectory information and trajectory explanations, enabling trajectory prediction evaluation.}
\vspace{-0.3em} 
\small 
\begin{tabular}{lccccccc}
\toprule 
\multirow{2}{*}{Model} & \multicolumn{3}{c}{Understanding} & \multicolumn{2}{c}{CoT} & Planning & Following \\
\cmidrule(lr){2-4} \cmidrule(lr){5-6} 
& Small   & Relationship & Abnor. Pred. & GPT    & Blue     & L2 (3s)  & L2 (3s)  \\
\midrule 
GPT-4o                & 64.2\%  & 63.5\%       & 72.8\%       & 0.55   & 0.125    & 2.63     & 2.58     \\
Qwen-2.5-VL-72B       & 75.8\%  & 74.9\%       & 81.5\%       & 0.72   & 0.188    & 1.94     & 1.89     \\
Our w/o CoT          & 86.5\%  & 85.7\%       & 93.2\%       & 0.83   & 0.218    & 1.58     & 1.53    \\
Our w/o Gen.      & 83.7\%  & 82.9\%       & 90.6\%       & 0.80   & 0.203    & 1.72     & 1.67       \\
Our               & \textbf{89.3\%} & \textbf{88.6\%} & \textbf{95.8\%} & \textbf{0.88} & \textbf{0.240} & \textbf{1.45} & \textbf{1.40}     \\
\bottomrule 
\end{tabular}
\label{tab:driving_performance}
\end{table*}

\begin{table*}[ht]
\setlength{\tabcolsep}{0.01\linewidth}
\caption{\textbf{End-to-end motion planning performance on the nuScenes~\cite{Caesar2019nuScenesAM} dataset.} 
Note that our model achieves a low collision rate, demonstrating its understanding of basic traffic rules via simple next-frame prediction. $^*$ represents only using the front camera as input. }
\centering
\resizebox{0.99\linewidth}{!}{
\begin{tabular}{l|lc|cccc|cccc}
\toprule
\multirow{2}{*}{Method} & \multirow{2}{*}{Input} & \multirow{2}{*}{Auxiliary Supervision} &
\multicolumn{4}{c|}{L2 (m) $\downarrow$} & 
\multicolumn{4}{c}{Collision Rate (\%) $\downarrow$} \\
&& & 1s & 2s & 3s & \cellcolor{gray!30}Avg. & 1s & 2s & 3s & \cellcolor{gray!30}Avg.  \\
\midrule
ST-P3~\cite{hu2022stp3} & Camera & Map \& Box \& Depth & 1.33 & 2.11 & 2.90 & \cellcolor{gray!30}2.11 & 0.23 & 0.62 & 1.27 & \cellcolor{gray!30}0.71  \\
UniAD~\cite{hu2023_uniad} & Camera & { \footnotesize Map \& Box \& Motion \& Tracklets \& Occ}  & {0.48} & {0.96} & {1.65} & \cellcolor{gray!30}{1.03} & {0.05} & {0.17} & {0.71} & \cellcolor{gray!30}{0.31}  \\
OccWorld~\cite{zheng2023occworld} & Camera & 3D-Occ & 0.52 & 1.27 & 2.41 & \cellcolor{gray!30}1.40 & 0.12 & 0.40 & 2.08 & \cellcolor{gray!30}0.87  \\
VAD-Tiny~\cite{jiang2023vad}  & Camera & Map \& Box \& Motion  & 0.60 & 1.23 & 2.06 & \cellcolor{gray!30}1.30 & 0.31 & 0.53 & 1.33 & \cellcolor{gray!30}0.72  \\
VAD-Base~\cite{jiang2023vad} & Camera & Map \& Box \& Motion & 0.54 & 1.15 & 1.98 & \cellcolor{gray!30}1.22 & 0.04 & 0.39 & 1.17 & \cellcolor{gray!30}0.53 \\
GenAD~\cite{zheng2024genad} & Camera & Map \& Box \& Motion & {0.36} & {0.83} & {1.55} & \cellcolor{gray!30}{0.91} & 0.06 & {0.23} & {1.00} & \cellcolor{gray!30}{0.43} \\
\midrule
Doe-1~\cite{doe} & Camera$^*$ & QA & 0.50 & 1.18 & 2.11 & \cellcolor{gray!30}1.26 & 0.04 & 0.37 & 1.19 & \cellcolor{gray!30}0.53  \\
Epona~\cite{Zhang2025EponaAD}& Camera$^*$ & None & 0.61 & 1.17  & 1.98 & \cellcolor{gray!30}1.25 & \textbf{0.01} &  0.22 & 0.85 & \cellcolor{gray!30}0.36  \\
UniUGP (Ours)& Camera$^*$ & QA & \textbf{0.58} & \textbf{1.14}  & \textbf{1.95} & \cellcolor{gray!30}\textbf{1.23} & \textbf{0.01} &  \textbf{0.19} & \textbf{0.81} & \cellcolor{gray!30}\textbf{0.33}  \\
\bottomrule
\end{tabular}%
}
\label{tab:plan_nusc}
\end{table*}

\subsection{Evaluation of Planning Ability} 

As shown in Table~\ref{tab:plan_nusc}, our model (Ours) achieves competitive performance under the setting of front camera input (Camera$^*$) and QA auxiliary supervision: it attains an average L2 distance of 1.23m and an average collision rate of 0.33\%, outperforming multiple comparative methods with similar input constraints. Specifically, compared to Doe-1~\cite{doe} with the same input and auxiliary supervision (Camera$^*$+QA), our model reduces the average L2 distance from 1.26m to 1.23m and the average collision rate from 0.53\% to 0.33\%. It also performs favorably against advanced methods like GenAD~\cite{zheng2024genad} (average L2: 0.91m, collision rate: 0.43\%) and UniAD~\cite{hu2023_uniad} (average L2: 1.03m, collision rate: 0.31\%) considering our more constrained input (only front camera vs. full camera suite). Additionally, our model surpasses Epona~\cite{Zhang2025EponaAD} (Camera$^*$+None, average L2: 1.25m, collision rate: 0.36\%) under similar input conditions, with lower average L2 distance and collision rate. These results demonstrate the effectiveness of our unified model’s capabilities in trajectory planning accuracy and driving safety even with a unified model.

\subsection{Evaluation of Generation Ability} 
\begin{figure*}[htbp]
    \centering
    \includegraphics[width=0.82\textwidth]{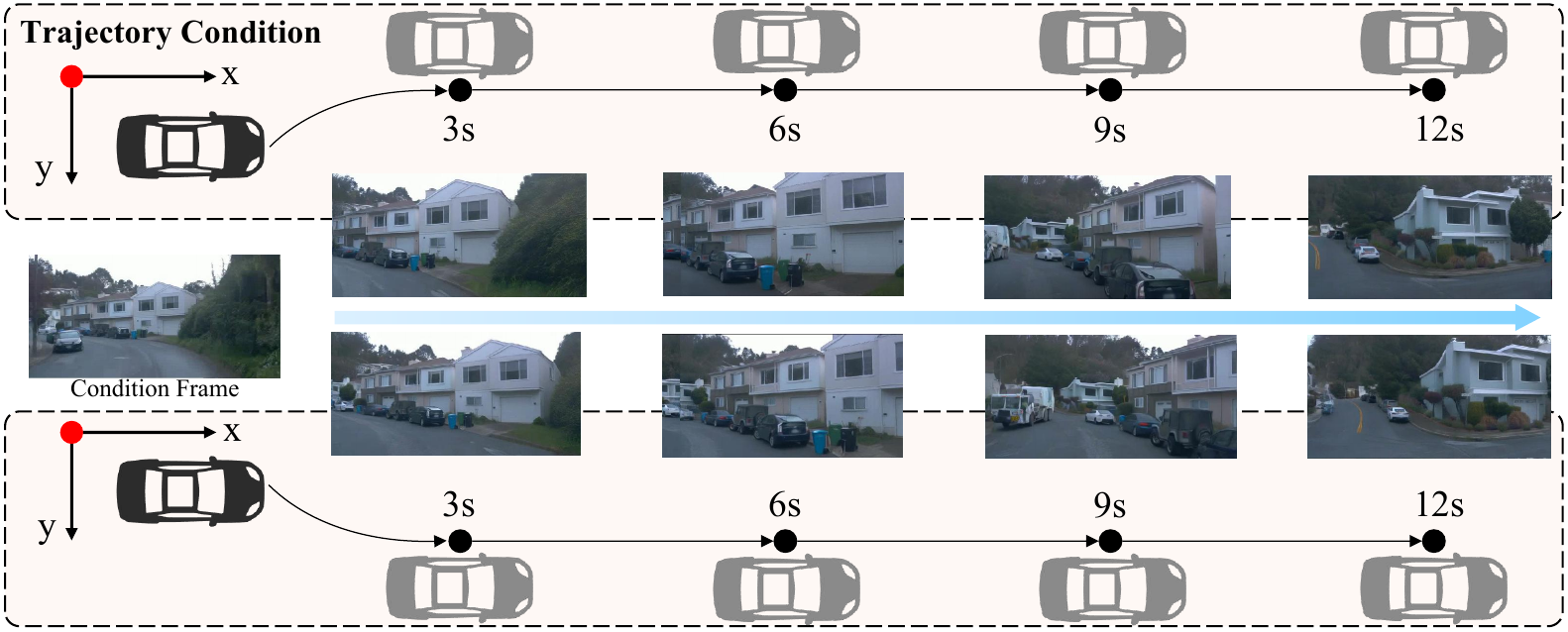}
    \caption{\textbf{Trajectory controllable generation visualization.} We control the generation of future frames of the video by modifying the trajectories fed into the generation model, which demonstrates the controllability of our generation experts.} 
    \label{fig:gen}
\end{figure*}

As shown in Table~\ref{tab:fid}, our method is evaluated under the same protocol as Epona~\cite{Zhang2025EponaAD} and FSDrive~\cite{zeng2025futuresightdrive}, achieving significant improvements in generation quality. This gain stems from the effective utilization of a pre-trained generative model, which enhances the model's ability to capture realistic scene dynamics and appearance. We have provided a trajectory-controllable visualization as shown in Fig.~\ref{fig:gen}, which demonstrates the controllability of our generated experts.

\begin{table*}[!t]
\centering
\caption{\textbf{Future Frames Generation Quality Comparison on the NuScenes Dataset.}}
\setlength{\tabcolsep}{0.3mm} 
\resizebox{\linewidth}{!}{
\begin{tabular}{l|ccccccc|c}
\toprule
\multirow{2}{*}{\textbf{Method}}& DriveDreamer~\cite{wang2023drivedreamer}&Drive-WM~\cite{wang2023driving}& GenAD~\cite{yang2024genad} & GEM~\cite{Hassan2024GEMAG} &Doe-1~\cite{doe}& Epona~\cite{Zhang2025EponaAD} &FSDrive~\cite{zeng2025futuresightdrive}& \modelname  \\
&[ECCV24]&[CVPR24]&[CVPR24] & [CVPR25]&[arXiv24]&[ICCV25] & [NeurIPS25] & Ours   \\ 
\midrule
\textbf{Type}&Diff&Diff&Diff&Diff&AR&Diff&AR &AR+Diff \\ 
\textbf{Res.}   & 128×192 & 192×384 & 256×448 & 576×1024& 384×672&576×1024 &128×192 &512×512\\ 
\midrule
\textbf{FID} $\downarrow$& 52.6& 15.8& 15.4&10.5 &15.9&7.5 &10.1 & \textbf{7.4}\\ 
\textbf{FVD} $\downarrow$& 452.0 & 122.7 & 184.0 & - & - &82.8 & - & \textbf{75.9}\\ 
\bottomrule
\end{tabular}
}
\label{tab:fid}
\end{table*}

\subsection{Results on DriveLM Dataset} 

As shown in Table~\ref{tab:vqa}, under the same evaluation protocol as FSDrive~\cite{Wang2024OmniDriveAH}, our model (\modelname) achieves superior performance on the DriveLM GVQA Benchmark compared to existing state-of-the-art methods. Our final score reaches 0.59, which is higher than FSDrive (0.57), OmniDrive~\cite{Wang2024OmniDriveAH} (0.56), SimpleLLM4AD~\cite{Zheng2024SimpleLLM4ADAE} (0.53), TrackingMeetsLMM~\cite{ishaq2025trackingmeets} (0.52), Cube-LLM~\cite{cube-llm} (0.50), and the DriveLM baseline~\cite{sima2023drivelm} (0.32). Across key metrics, our accuracy is 0.74, outperforming all comparative methods; BLEU is 0.78 and ROUGE is 0.76, both leading FSDrive (0.76, 0.74) and other SOTA models; the match metric is 0.41, which is also higher than existing methods. While our ChatGPT score (0.64) and CIDEr score (0.19) are competitive among the methods, the overall leading performance across core evaluation dimensions demonstrates the advantage of our method over existing state-of-the-art approaches in scene understanding and language interaction capabilities.

\begin{table*}[!t]
\footnotesize
\begin{center}
\setlength{\tabcolsep}{1.5mm}
\caption{\textbf{Results on DriveLM GVQA Benchmark.}}
\begin{tabular}{l|cccccc|c}
\toprule
\textbf{Method}& \textbf{Acc.} $\uparrow$ &\textbf{GPT} $\uparrow$&\textbf{BLEU\_1} $\uparrow$ &\textbf{ROUGE\_L} $\uparrow$&\textbf{CIDEr} $\uparrow$&\textbf{Match} $\uparrow$&\cellcolor{gray!30}\textbf{Final Score} $\uparrow$ \\
\midrule
DriveLM baseline~\pub{ECCV24}~\cite{sima2023drivelm}&0.00&0.65&0.05 &0.08  &0.10&0.28&\cellcolor{gray!30}0.32 \\
Cube-LLM~\pub{ICLR25}~\cite{cube-llm}&0.39&\textbf{0.89} &0.16 &0.20 &\textbf{0.31} &0.36&\cellcolor{gray!30}0.50 \\
TrackingMeetsLMM~\pub{arxiv25}~\cite{ishaq2025trackingmeets}& 0.60& 0.58& 0.72&  0.72& 0.04& 0.36 &\cellcolor{gray!30}0.52  \\
SimpleLLM4AD~\pub{arxiv24}~\cite{Zheng2024SimpleLLM4ADAE}&0.66& 0.57 &0.76& 0.73 &0.15&0.35& \cellcolor{gray!30}0.53  \\
OminiDrive~\pub{CVPR25}~\cite{Wang2024OmniDriveAH}&0.70&0.65&0.52 &0.73 &0.13&0.37&\cellcolor{gray!30}0.56\\
FSDrive~\pub{NeurIPS25}~\cite{Wang2024OmniDriveAH} &0.72&0.63&0.76 &0.74 &0.17&0.39&\cellcolor{gray!30}0.57\\ 
\midrule
\modelname (Ours) &\textbf{0.74}&0.64&\textbf{0.78} &\textbf{0.76} &0.19&\textbf{0.41}&\cellcolor{gray!30}\textbf{0.59}\\ 
\bottomrule
\end{tabular}
\label{tab:vqa}
\end{center}
\vspace{-3mm}
\end{table*}




\section{Conclusion}

We propose \modelname, a unified Understanding-Generation-Planning framework addressing critical challenges in temporal dynamics and long-tail generalization. Leveraging a hybrid expert architecture with pre-trained VLMs and video generation models, \modelname produces interpretable reasoning, physically consistent trajectories, and coherent future videos from multimodal inputs. Our four-stage training strategy progressively aligns these capabilities across diverse datasets. Extensive experiments demonstrate state-of-the-art performance and robust generalization, establishing a strong foundation for future autonomous driving research.

\phantomsection
\section*{Contributions}
\label{sec:contribution}
\textbf{Authors:} Hao Lu$^{1,2,*, \dagger}$, Ziyang Liu$^{2, *}$, Guangfeng Jiang$^{3}$, Yuanfei Luo$^{2, \dagger}$, Sheng Chen$^{2}$, Yangang Zhang$^{2}$, Ying-Cong Chen$^{1, \S}$

\textbf{Affiliations:} $^{1}$HKUST-GZ, $^{2}$ByteDance Seed, $^{3}$Independent Researcher

$^{*}$ Co-first authors \\
$^{\dagger}$ Project leads \\
$^{\S}$ Corresponding author. \\
Note: Hao LU work done at ByteDance Seed during internship.

%% file: sections/appendix.tex
\section{Dataset}
To comprehensively evaluate and enhance the multimodal world modeling capability required for end-to-end autonomous driving, we reconstruct heterogeneous open-source driving datasets into a unified framework aligned with four essential cognitive competencies. Existing benchmarks mainly emphasize structured urban scenes or closed-loop simulation metrics, which fail to capture a model’s robustness in open-world, long-tail, and low-probability-but-high-risk scenarios. In contrast, our objective is to measure how well a model can understand, reason, and act under diverse real-world conditions. 

\begin{figure*}[htbp]
    \centering
    \begin{subfigure}[b]{0.95\textwidth}
        \centering
        \includegraphics[width=\textwidth]{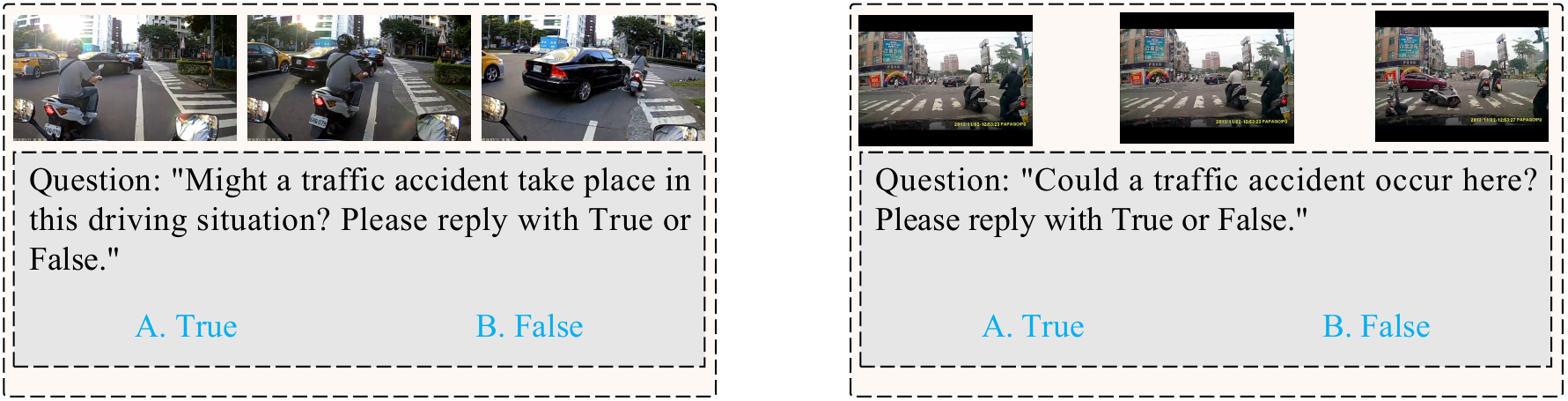} 
    \end{subfigure}
    \vspace{0.5cm} 
    \begin{subfigure}[b]{0.95\textwidth}
        \centering
        \includegraphics[width=\textwidth]{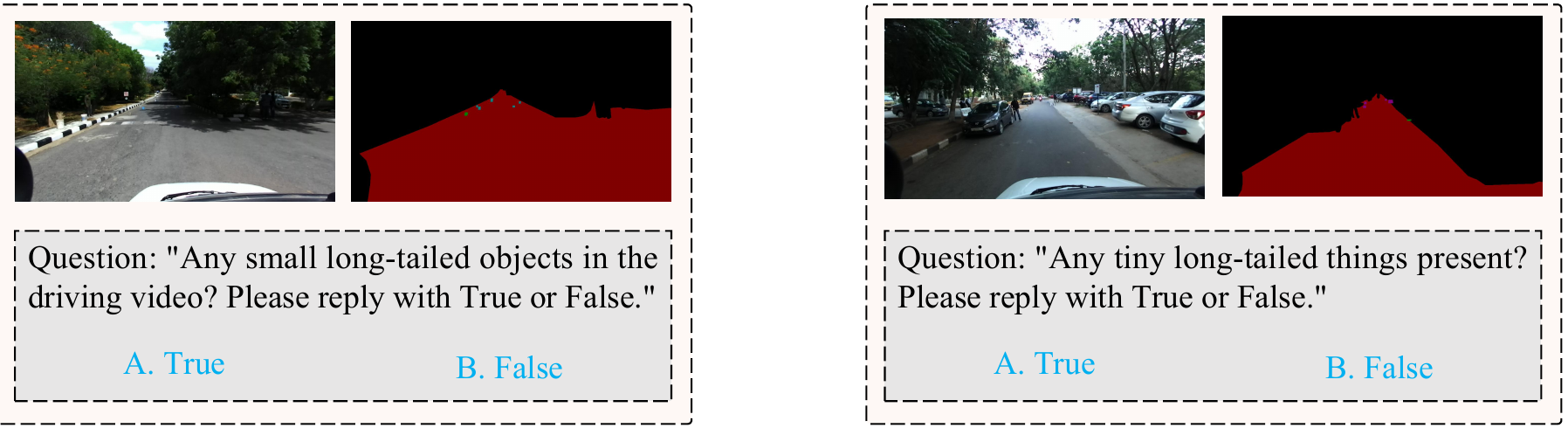} 
    \end{subfigure}
    \begin{subfigure}[b]{0.95\textwidth}
        \centering
        \includegraphics[width=\textwidth]{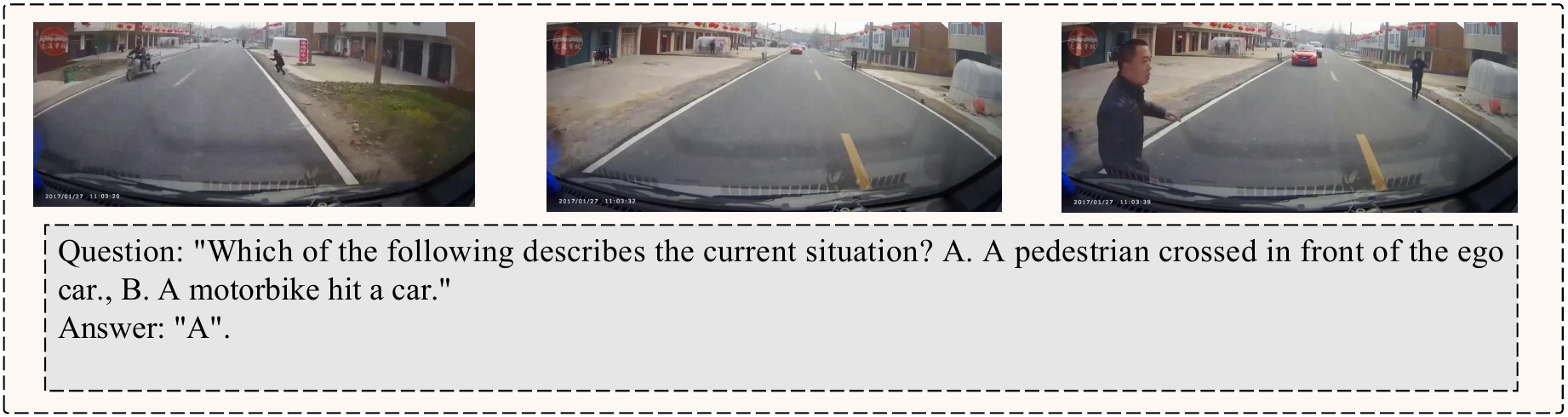} 
    \end{subfigure}
    \caption{Long-tail perception and understanding of questions and answers. }
    \label{fig:pu}
\end{figure*}

To this end, we integrate datasets covering road abnormalities and traffic accident anticipation (e.g., Lost and Found, StreetHazards, DADA-2000, and Anticipating Accidents in Dashcam Videos) together with the Waymo dataset, which contains densely interactive and ambiguous safety-critical events. We reorganize these data according to four key task dimensions: \textbf{(1) Perception and Understanding}, assessing visual semantics and contextual risk awareness; \textbf{(2) Causal CoT Reasoning}, explaining the underlying causes behind motion intentions; \textbf{(3) Planning and Decision-Making}, supervising physically feasible future trajectories; and \textbf{(4) Instruction Following}, evaluating whether control actions align with high-level navigation commands. This integrated categorization supports unified training and interpretable evaluation of cognitive, generative, and control capabilities.

\subsection{Perception and Understanding:} We convert scene annotations from anomaly and hazard datasets into true/false and multiple-choice QA items that evaluate not only basic semantic comprehension, but also driving commonsense, corner-case interpretation, and the recognition of previously unseen or rare object categories. This formulation ensures that the model can identify what is present in the scene, understand why it is safety-relevant, and generalize to atypical or long-tail conditions critical for real-world driving. Specific examples are shown in Fig~\ref{fig:pu}.

\textbf{Small long-tailed object.}
We collected multiple datasets that contained small elongated objects. We determine whether there are small tail objects based on the segmentation map labels provided by the dataset, and thereby construct a judgment question. We designed many random questions to enhance the generalization ability of the model's question answering. The questioning format is as shown in List~\ref{lst:question_templates}:
\begin{lstlisting}[caption=Small long-tailed object, label=lst:question_templates]
## Prompt = Question + "Please reply with True or False."
# Question:
"Any small long-tailed objects in the driving video?",
"Are there tiny long-tailed items in the driving clip?",
"Does the driving video have small long-tailed things?",
"Small long-tailed objects present in the driving footage?",
"Are there small long-tailed objects in the driving video?",
"Any small long-tailed objects?",
"Are there small items with long tails?",
"Exist small objects with long tails?",
"Any tiny long-tailed things present?",
\end{lstlisting}

\textbf{Long-tailed accident prediction.}
We collected videos of abnormal traffic accidents. The dataset classifies whether the video is abnormal based on whether it is abnormal as indicated by the dataset and the specific timestamp. We have designed various questions to enhance generalization. The questioning method is as shown in List~\ref{lst:lta}:

\begin{lstlisting}[caption=Long-tailed accident prediction, label=lst:lta]
## Prompt = Question + "Please reply with True or False."
# Question:
"Has an abnormal accident occurred here?", 
"Could there be any traffic hazards here?",
"Could a traffic accident occur here?",
"Are there any traffic risks or hidden hazards here that could lead to accidents?",
"Was there an unusual incident or accident here just now?",
"Might a traffic accident take place in this driving situation?",
"Is there a possibility that a traffic accident will happen here?"
\end{lstlisting}

\textbf{Long-tailed accident relationship.}
We collected video footage of abnormal traffic accidents, along with annotations indicating the abnormal entities involved. Based on these annotations, we designed multiple-choice questions. The questioning method is as shown in the list~\ref{lst:mcc}:

\begin{lstlisting}[caption=Long-tailed accident relationship, label=lst:mcc]
INPUT: correct_answer (str), candidate_pool (List[str])
OUTPUT: question (str), options_dict (Dict[str, str]), correct_label (str), correct_answer (str)

# Step 1: Filter distractor candidates (exclude correct answer)   
distractor_candidates = [opt for opt in candidate_pool if opt ≠ correct_answer]  

# Step 2: Randomly select question type (2-option / 4-option)
option_count = RANDOM_CHOICE([2, 4])
distractor_needed = option_count - 1

# Step 3: Select distractors (supplement if insufficient)
if len(distractor_candidates) ≥ distractor_needed: 
    selected_distractors = RANDOM_SAMPLE(distractor_candidates, distractor_needed)
else:
    selected_distractors = distractor_candidates + RANDOM_CHOICES(distractor_candidates, k=distractor_needed - len(distractor_candidates))

# Step 4: Combine and shuffle options
all_options = [correct_answer] + selected_distractors
SHUFFLE(all_options)

# Step 5: Map to option labels (A/B/C/D)
option_letters = ['A', 'B', 'C', 'D'][:option_count]
options_dict = dict(zip(option_letters, all_options))  

# Step 6: Locate correct label
correct_label = next(letter for letter, opt in options_dict.items() if opt == correct_answer) 

# Step 7: Generate question with specified template
question = "Which of the following describes the current situation? " + ', '.join([f"{letter}. {opt}" for letter, opt in options_dict.items()])  

RETURN question, options_dict, correct_label, correct_answer
\end{lstlisting}

\begin{figure*}[htbp]
    \centering
    \begin{subfigure}[b]{0.95\textwidth}
        \centering
        \includegraphics[width=\textwidth]{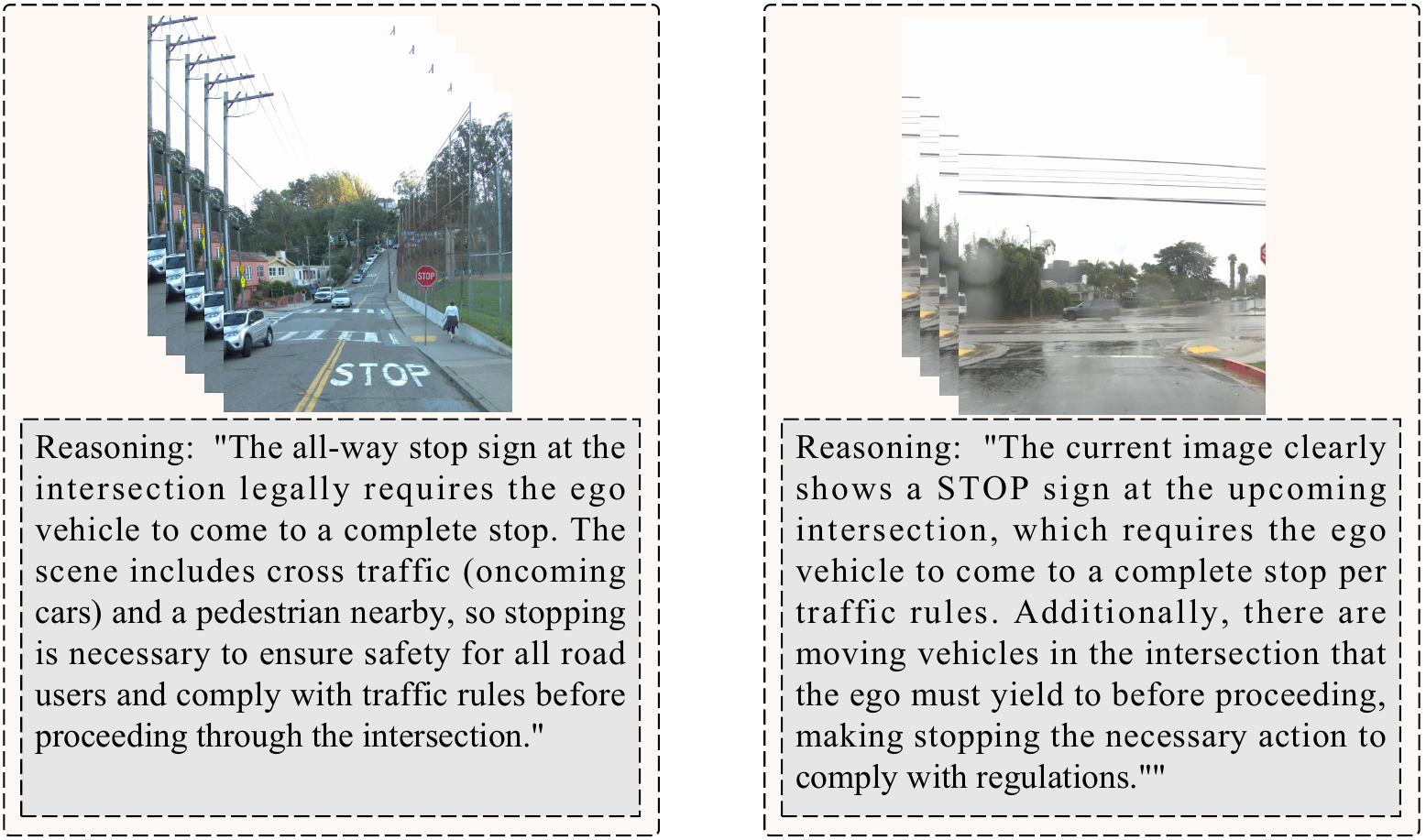} 
    \end{subfigure}
    \caption{Future Trajectories-Based CoT Reasoning. }
    \label{fig:cot}
\end{figure*}

\subsection{Causal CoT Reasoning:} For sequences where the future driving outcome is observable, we construct QA pairs in which the answer is a structured multi-step chain of thought that explains how the final driving decision is formed. During dataset construction, we utilize both the future image sequence and the ground-truth ego trajectory to ensure that the reasoning process is strictly aligned with the actual physical outcome. The reasoning is required to describe the scene context, identify key interactive agents, infer their potential intentions, and justify the final driving action that leads to the observed future behavior. This design encourages the model to learn reasoning that is predictive and causally grounded in realistic driving dynamics, instead of producing descriptive or speculative explanations after the fact. As a result, the model is encouraged to understand why a particular action is necessary for safety rather than merely recognizing what action is taken. Specific examples are shown in Fig~\ref{fig:cot}.
CoT reasoning based on future trajectories is presented in List~\ref{lst:gpt_prompt}:

\begin{lstlisting}[caption=Future Trajectories-Based CoT Reasoning, label=lst:gpt_prompt, breaklines=true]
## The prompts for generating CoT:

You are an autonomous driving planning expert. Your task is to analyze a driving scene using information provided from the images of front camera views, past waypoints, past velocity, past acceleration, current velocity, current acceleration, future waypoints, and the intended motion of the ego vehicle. You must reason through the scene based on visual and temporal data and produce a structured decision output. I am giving your future decisions to help you confirm that your reasoning is correct, but your reasoning process should not rely on the future trajectory as the basis for your reasoning. Your reasoning can only make predictions about future decisions based on the current image and historical trajectory.
 
---
**Input:**

- **Past waypoints**: %%<past_waypoints>%%  
- **Past velocity**: %%<past_velocity>%%  
- **Past acceleration**: %%<past_acceleration>%%  
- **Future waypoints**: %%<future_waypoints>%%  

---

**Coordinate and Temporal System Explanation:**

> All motion and state data are expressed in the ego local coordinate system:
> - +x = forward direction  
> - +y = left direction  
> - The origin (0, 0) is located at the center of the ego vehicle

> - Units:  
>   - Waypoints: meters  
>   - Velocity: meters/second  
>   - Acceleration: meters/second  

> - Sampling frequency: 4Hz (every 0.25s)

> - Past arrays (`past_waypoints`, `past_velocity`, `past_acceleration`) are each 16 time steps:  
>   - Format: `[ [x0, y0], [x1, y1], ..., [x15, y15] ]`  
>   - Indexed `i = 0` to `15`, with  
>     - `i = 0` -> `t = -3.75s`,  
>     - `i = 15` -> `t = 0.0s` (current frame)

> - `Future_waypoints` are 20 time steps:  
>   - Format: `[ [x0, y0], [x1, y1], ..., [x19, y19] ]`  
>   - Indexed `i = 0` to `19`, with  
>     - `i = 0` -> `t = 0.25s`,  
>     - `i = 19` -> `t = 5.0s`

---

Your task: Based on the provided visual and motion data, complete the following four detailed Chain-of-Thought steps. I am giving your future decisions to help you confirm that your reasoning is correct, but your reasoning process should not rely on the future trajectory as the basis for your reasoning. Your reasoning can only make predictions about future decisions based on the current image and historical trajectory.

---

**Step 1: Scene Analysis**
Describe the overall traffic scene, including traffic lights, road geometry, lane markings, signs, crosswalks, construction areas, and any temporary road structures.

**Step 2: Key Object Identification**
Identify up to 3 important or rare objects that are relevant for planning, and also provide an assessment of their impact on driving, such as:
- Traffic signal light (if visible): describe its state (e.g., green, red, yellow, blinking, occluded), relative position, and meaning for ego vehicle
- Dynamic agents: describe the various attributes of each agent, its impact on the vehicle itself, and its significance in the planning of the vehicle entity.
- Rare objects (if visible):  e.g., traffic enforcers, construction workers, police officers, emergency responders, traffic cones, road barriers, parked delivery trucks, fallen trees, construction machinery, missing lane lines, misaligned curbs, temporary detours, unusual entry points

**Step 3: Intention Inference**
Based on input images (both current and future images), the intention of the ego vehicle, the past state of the ego vehicle (velocity, acceleration, waypoints), and the future waypoints of the ego vehicle, the possible future movement of the recognized object (for example, crossing the road, merging into the lane) is inferred. And how do these objects affect the future driving of ego vehicle.

**Step 4: Action Reason**
- action : Issue a command request for the future trajectory. For example, turn left and slow down in the future. Then this part should be filled in as "Please slow down and turn left.". This part should be as brief as possible.
- reason : Based solely on the current image and historical trajectory, provide a detailed Chain-of-Thought for this decision using step-by-step reasoning.
Please note that the "reason" should not involve any quantitative analysis. Provide a reasonable and necessary process for the decision-making reason.
I am giving your future decisions to help you confirm that your reasoning is correct, but your reasoning process should not rely on the future trajectory as the basis for your reasoning. Your reasoning can only make predictions about future decisions based on the current image and historical trajectory.
---

**Output Format:**
Your output format should be a json object with the following structure:
```json
{
  "scene_analysis": "...",
  "key_object": "...", 
  "intention_inference": "...",
  "action_decision": {
    "action": "...",
    "reason": "..."
  }
}
\end{lstlisting}

\subsection{Planning and Instruction Following} 

For planning task, we directly provide the model with a question that describes the current and historical driving context, and the model is required to predict the future trajectory of the ego vehicle for the next several seconds. The answer is represented as a sequence of K future trajectory points in the ego coordinate frame. The model may output this trajectory either directly based on its internal world understanding or after producing a CoT reasoning sequence when reasoning is beneficial for disambiguating complex interactions. This formulation ensures that trajectory prediction is guided by both contextual scene understanding and physically grounded motion patterns, enabling safe, smooth, and interpretable future driving behavior. 

To enable instruction-conditioned trajectory planning, we derive high-level navigation commands such as going straight, turning left, or turning right from the geometric properties of the ground-truth future trajectory. These commands are then incorporated into the QA pairs so that the model must generate a trajectory that is consistent with the given navigation intent. During training, the model learns to align the predicted motion sequence with the provided instruction, establishing a direct correspondence between semantic driving goals and trajectory generation. During evaluation, this setup allows us to assess whether the model can correctly adjust its predicted future motion according to the specified high-level driving intent, ensuring that the resulting planning behavior remains both controllable and interpretable in downstream driving scenes.

\section{More Cases}

\begin{figure*}[htbp]
    \centering
    \begin{subfigure}[b]{0.95\textwidth}
        \centering
        \includegraphics[width=\textwidth]{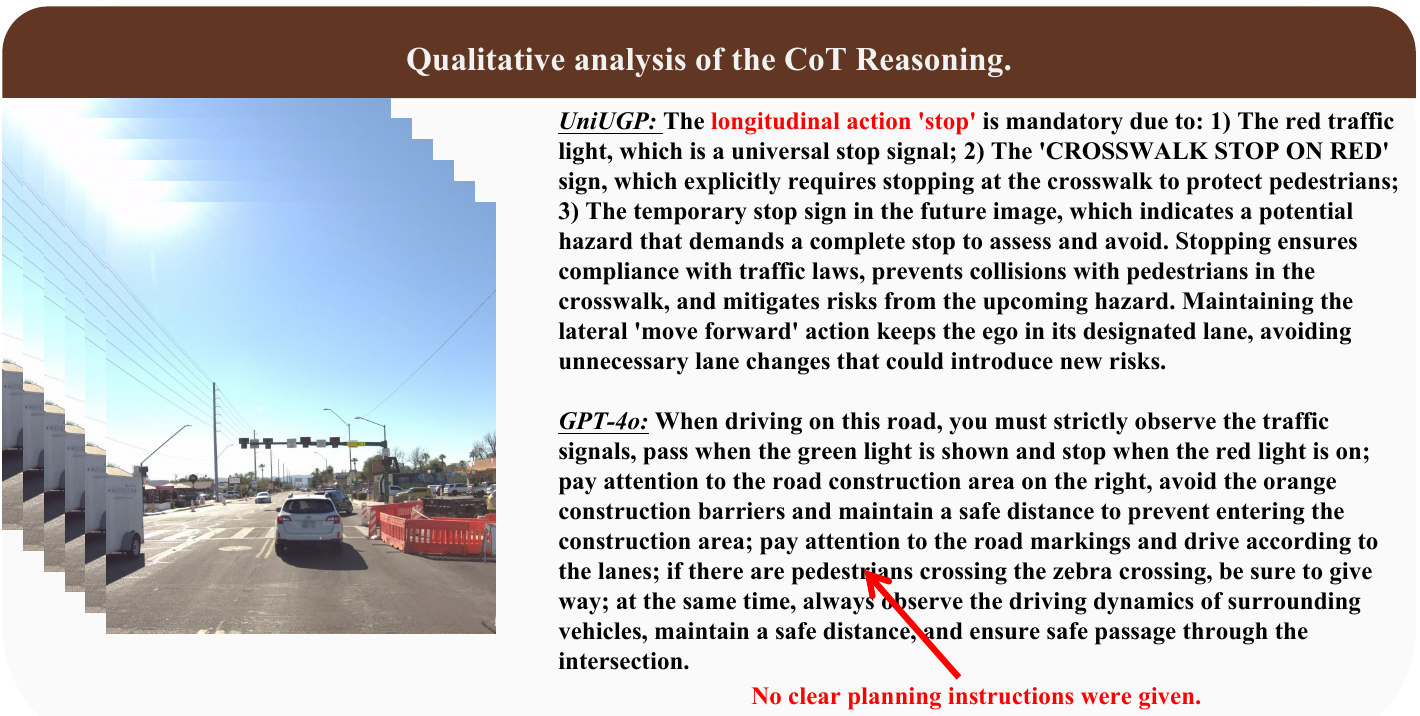} 
        \label{subfig:cot1}
    \end{subfigure}
    \vspace{0.5cm} 
    \begin{subfigure}[b]{0.95\textwidth}
        \centering
        \includegraphics[width=\textwidth]{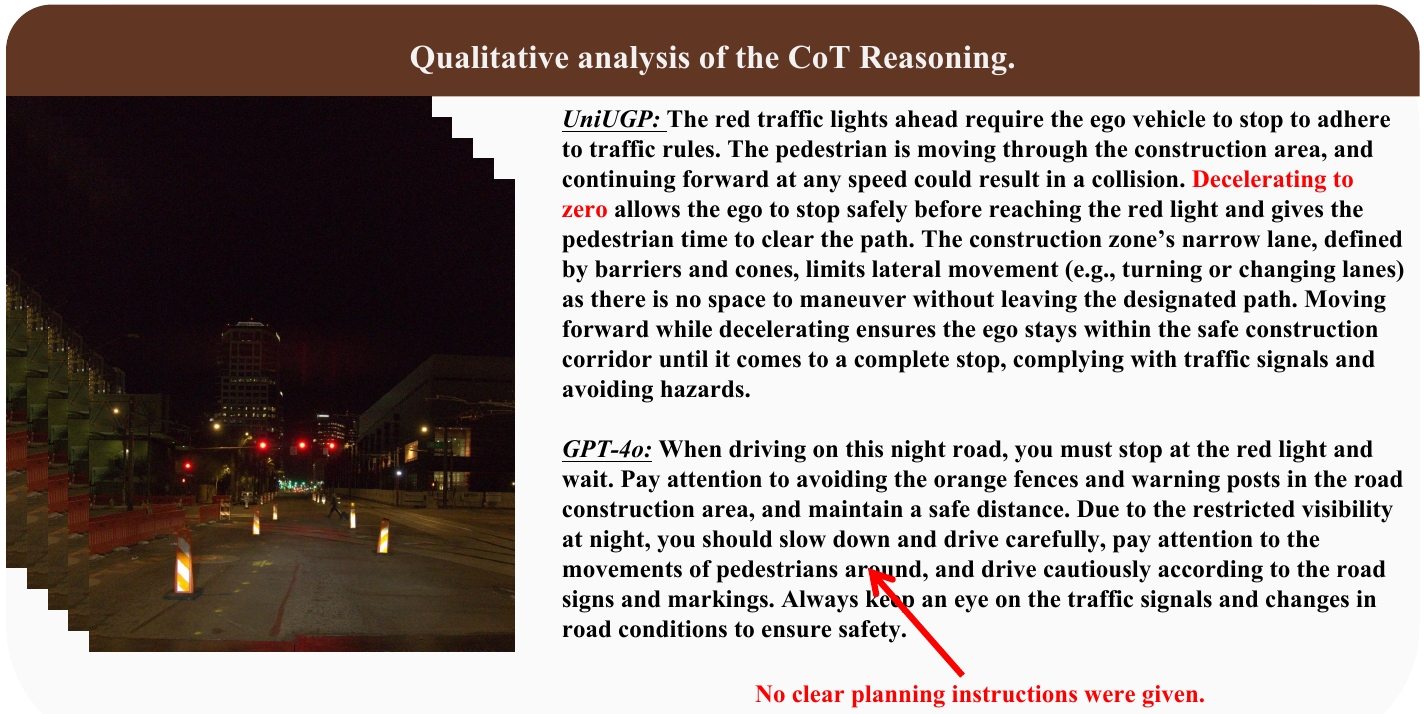} 
        \label{subfig:cot2}
    \end{subfigure}
    \caption{\textbf{Comparison of CoT in our method versus GPT4o.} Our approach provides more specific planning results, while the general large model does not offer sufficiently detailed planning outcomes.}
    \label{fig:method1}
\end{figure*}

\begin{figure*}[htbp]
    \centering
    \begin{subfigure}[b]{0.61\textwidth}
        \centering
        \includegraphics[width=\textwidth]{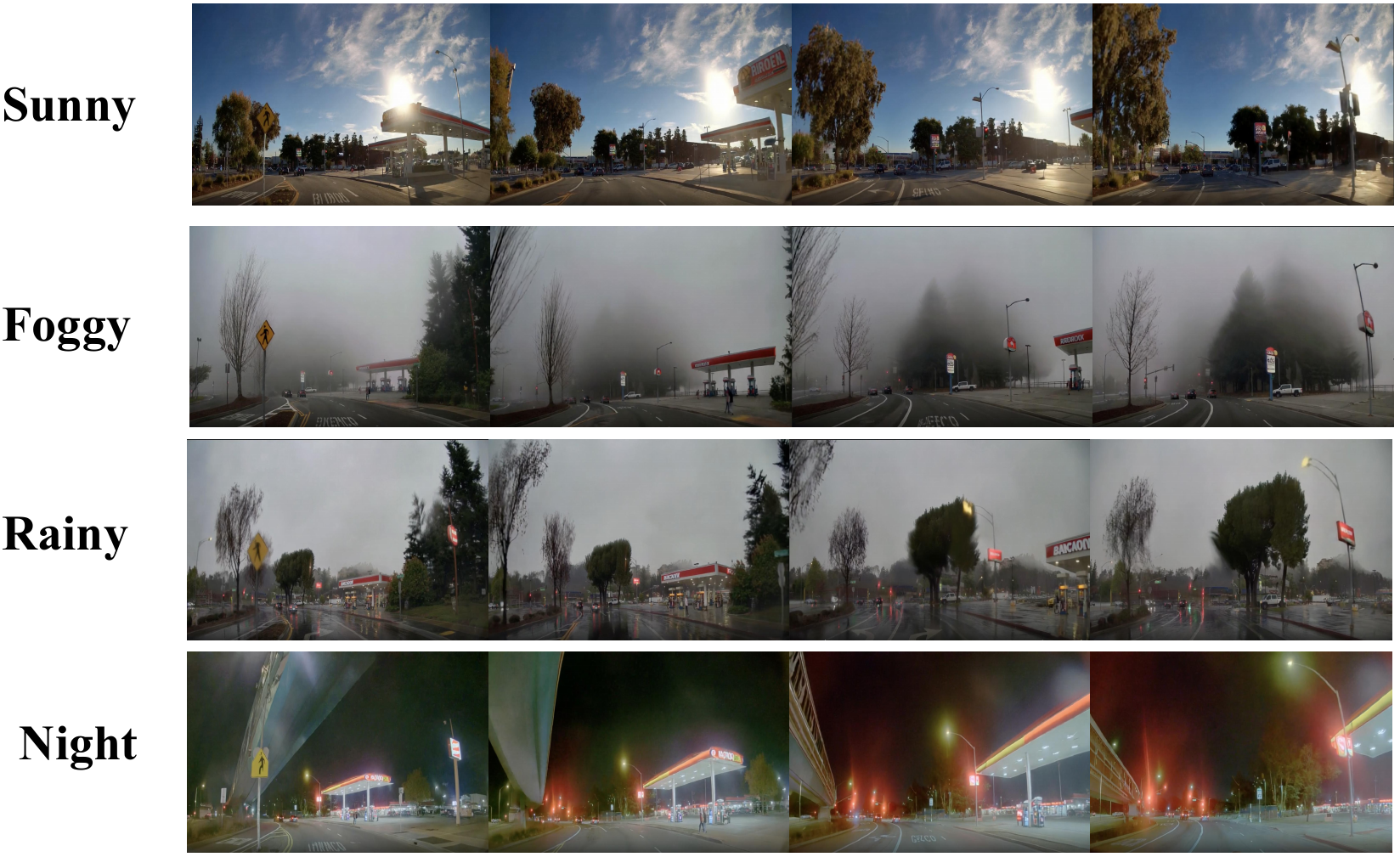} 
        \label{subfig:gen1}
    \end{subfigure}
    \vspace{0.5cm} 
    \begin{subfigure}[b]{0.6\textwidth}
        \centering
        \includegraphics[width=\textwidth]{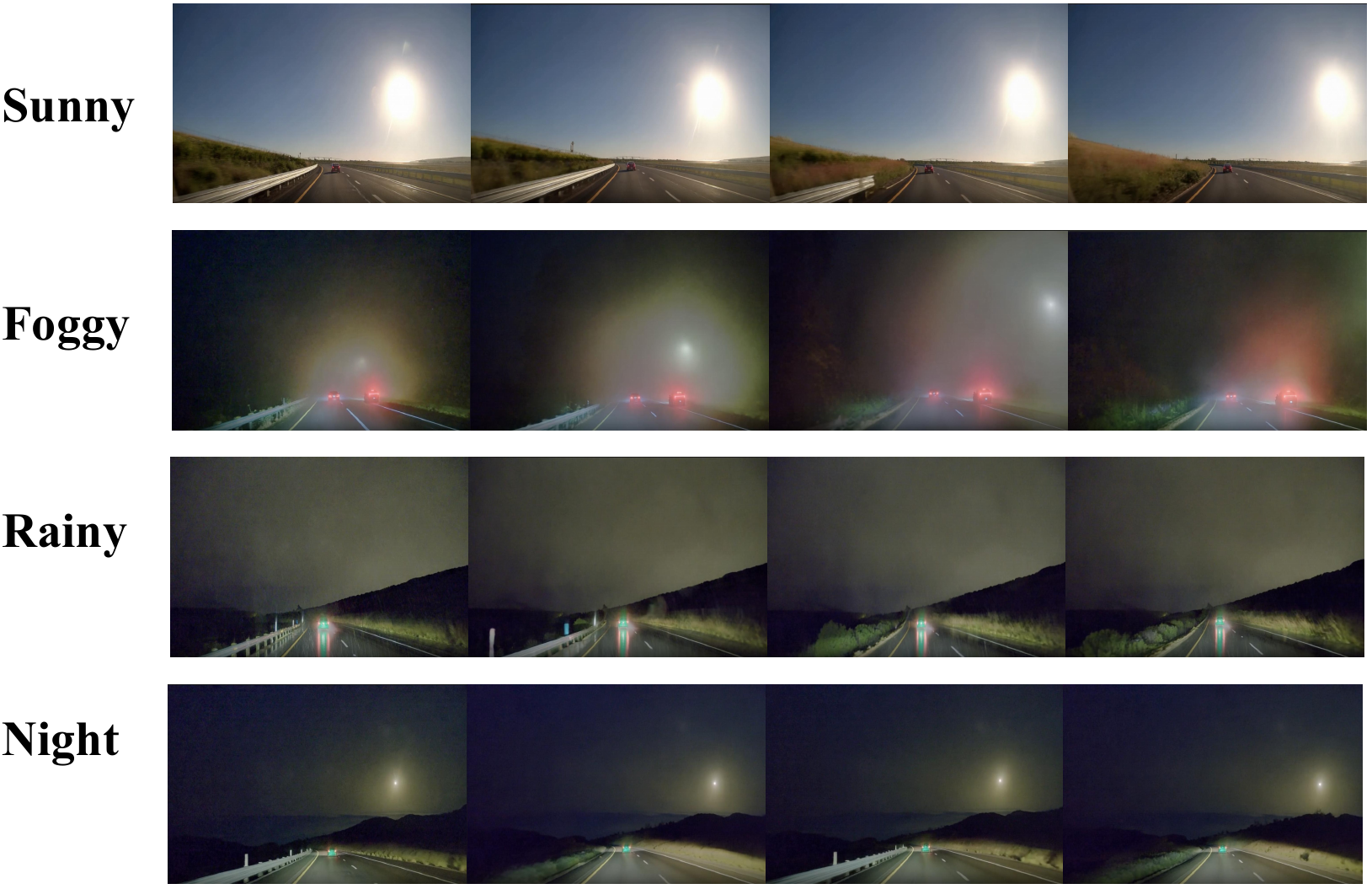} 
        \label{subfig:gen2}
    \end{subfigure}
    \caption{\textbf{Video generation visualization with controllable weather conditions.} Our model can generate videos of different weather conditions, which proves the efficiency of our generation model. Please zoom in to the best view.}
    \label{fig:method2}
\end{figure*}

\begin{figure*}[htbp]
    \centering
    \begin{subfigure}[b]{0.8\textwidth}
        \centering
        \includegraphics[width=\textwidth]{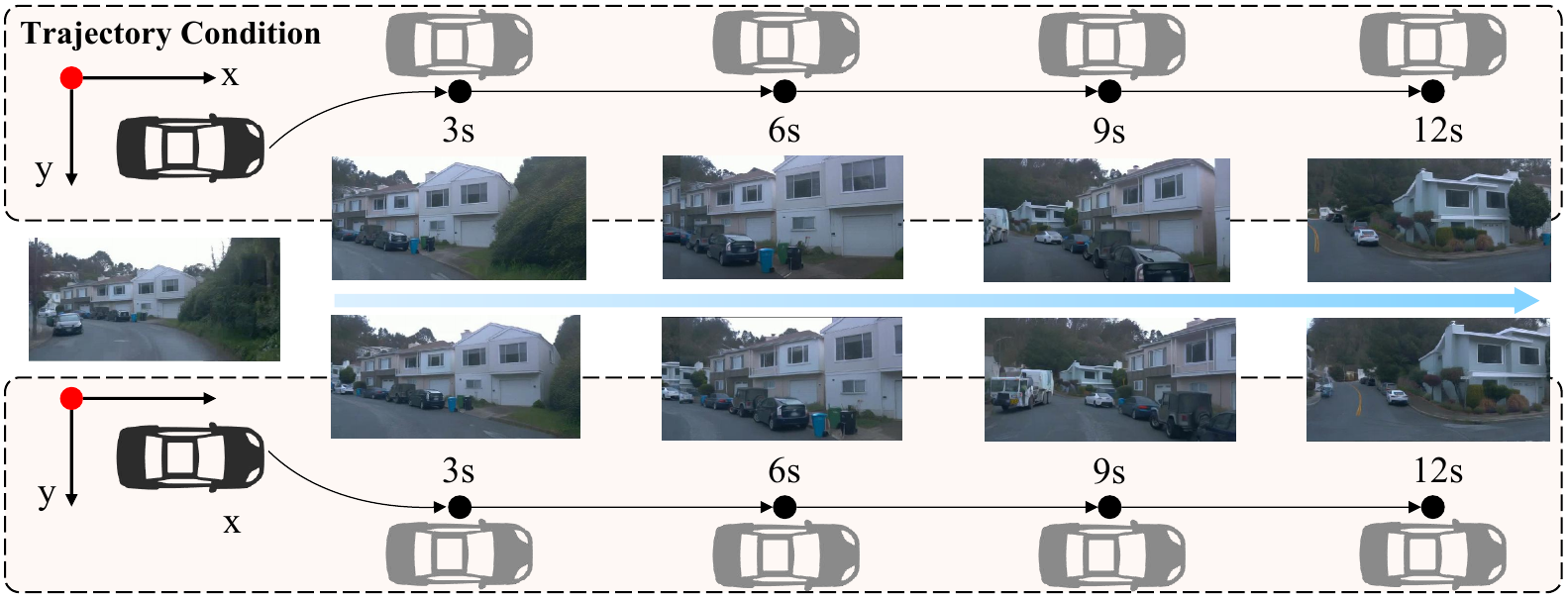} %
        \label{subfig:t1}
    \end{subfigure}
    \begin{subfigure}[b]{0.8\textwidth}
        \centering
        \includegraphics[width=\textwidth]{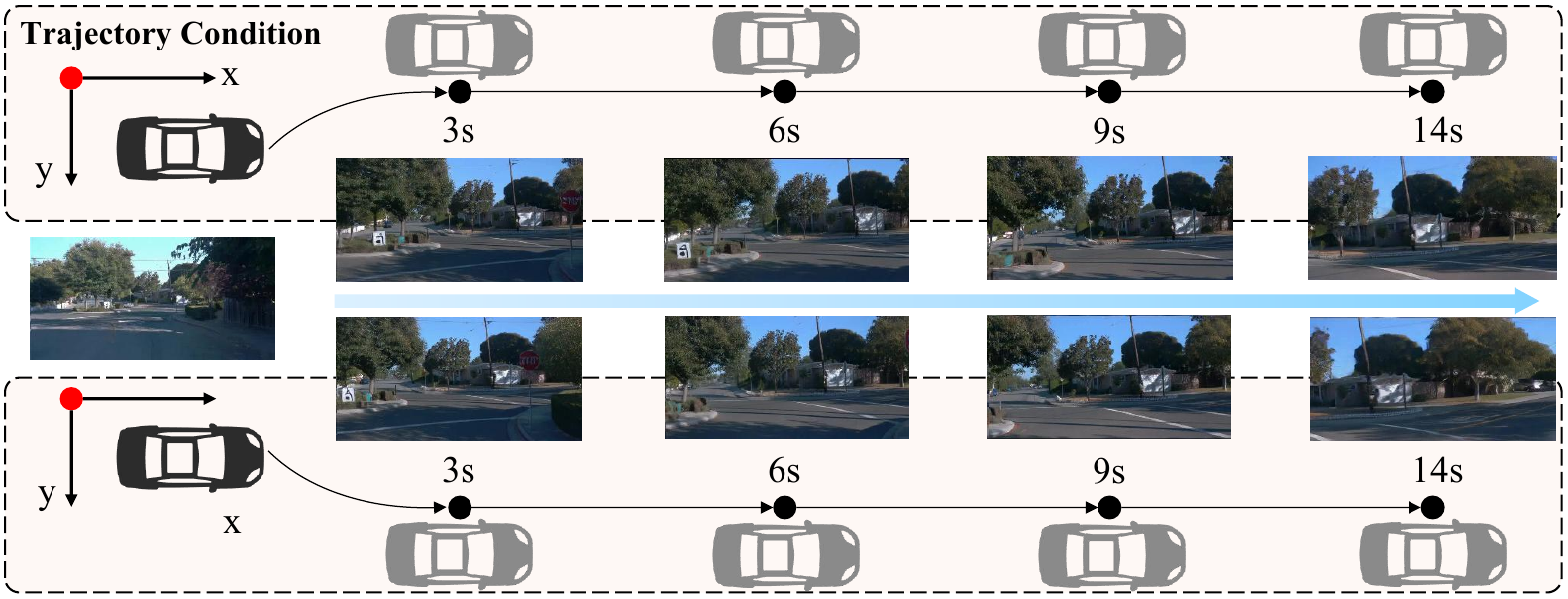}
        \label{subfig:t2}
    \end{subfigure}
    \begin{subfigure}[b]{0.8\textwidth}
    \centering
    \includegraphics[width=\textwidth]{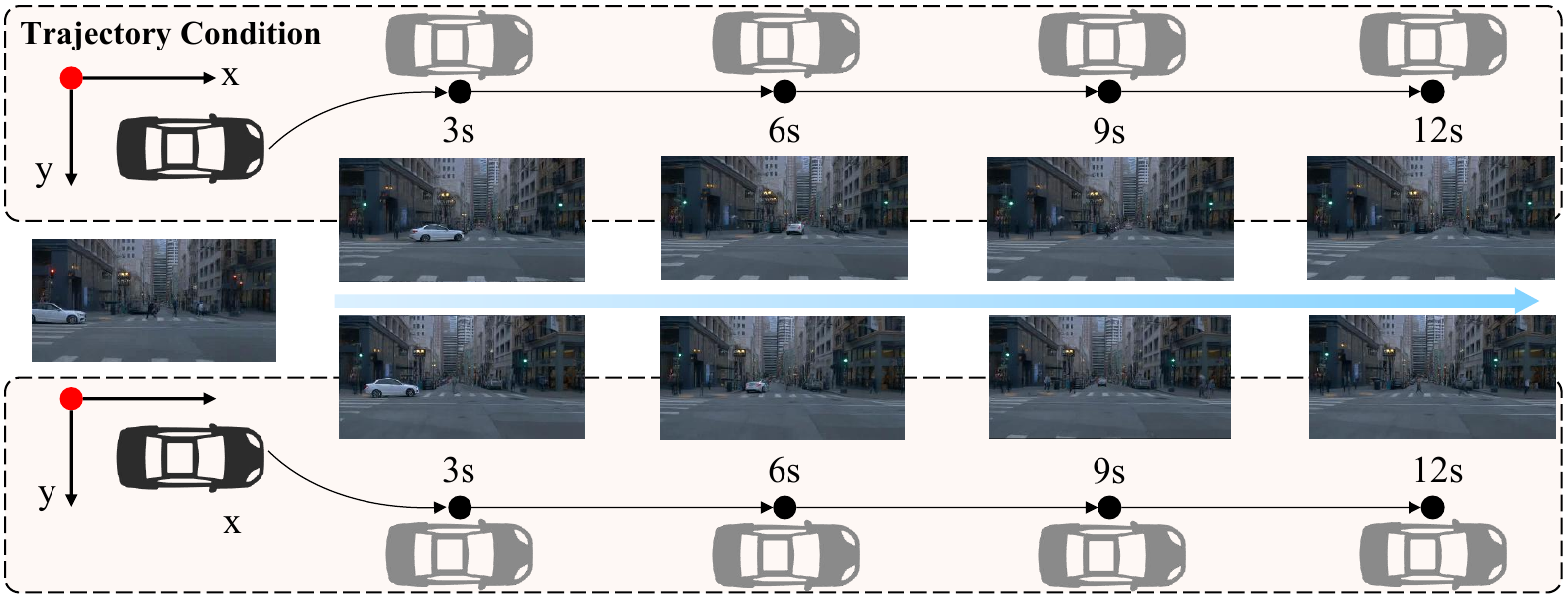} 
    \label{subfig:t3}
    \end{subfigure}
    \caption{\textbf{Video generation visualization with controllable trajectory conditions.} Our model can generate videos of different trajectory conditions, which proves the efficiency of our generation model. Please zoom in to the best view.}
    \label{fig:method3}
\end{figure*}

To better demonstrate the effectiveness of our method. We have provided even more examples. We compare the understanding and reasoning capabilities of the most advanced GPT4o in scenarios as shown in Fig.~\ref{fig:method1}. Our approach can provide more specific suggestions for planning. We have further provided visualizations for trajectory and weather control generation as shown in Fig.~\ref{fig:method2} and Fig.~\ref{fig:method3}. This proves the effectiveness of our method's generation capability.

\section{Limitation and Future Directions}
\label{sec:limitation_future}

Despite the promising performance of UniUGP in unifying scene understanding, future video generation, and trajectory planning for autonomous driving, it still has several limitations that point to valuable future research directions.

\subsection{Limitations}

First, while UniUGP uses over 10 diverse AD datasets to cover common and long-tail scenarios, its generalization to extreme rare events (e.g., unprecedented weather, novel obstacles) is constrained by training data coverage—critical for safety-critical systems. Second, the hybrid expert architecture’s computational efficiency is problematic: the generation expert, though useful for visual causal validation, demands excessive resources and must be disabled on resource-constrained mobile platforms to ensure real-time performance. Third, linguistic reasoning and physical dynamics alignment, though improved via multi-term loss functions, is suboptimal; in complex interactive scenarios (e.g., ambiguous pedestrian-vehicle interaction), chain-of-thought (CoT) reasoning may not tightly couple with physically consistent trajectory generation, causing minor interpretability-action inconsistencies. Fourth, the four-stage training strategy relies on fixed dataset proportions in the final fusion stage, failing to dynamically adapt to different datasets’ complementary strengths and limiting task synergy.

\subsection{Future Directions}

To address these limitations, we propose targeted directions: enhance generalization to extreme long-tail scenarios via high-fidelity synthetic data generation (e.g., world models + generative AI) and few-shot/zero-shot learning; optimize model efficiency through lightweight generation expert designs (e.g., knowledge distillation, sparse activation) and reduced multi-expert redundant computations. Deepen multimodal alignment with cross-modal contrastive learning and hierarchical fusion mechanisms that adjust expert weights by scene complexity; reduce labeled data dependence via self-supervised signals (e.g., unsupervised video causal reasoning) and enable incremental adaptation with continual learning to avoid catastrophic forgetting. Extend interaction capabilities to dynamic real-time feedback (e.g., mid-task voice commands) and multi-agent reasoning for complex traffic interactions; integrate UniUGP into closed-loop systems for real-world testing, establishing a performance-refinement feedback loop to boost safety and robustness. These efforts will evolve UniUGP into a more practical framework bridging laboratory performance and real-world deployment.

